\documentclass[letterpaper, 10 pt, conference]{ieeeconf}  

\IEEEoverridecommandlockouts                              
\usepackage[utf8]{inputenc} 
\usepackage[T1]{fontenc}    
\usepackage{hyperref}       
\usepackage{url}            
\usepackage{booktabs}       
\usepackage{amsfonts}       
\usepackage{amsmath}
\usepackage{nicefrac}       
\usepackage{microtype}      
\usepackage{xcolor}         
\usepackage{mwe}
\usepackage{siunitx}
\usepackage{graphicx}
\usepackage{subcaption}
\usepackage{caption}
\usepackage{layouts}
\usepackage{multirow}
\usepackage{makecell}
\usepackage{cleveref}

\title{\LARGE \bf
Learning Action-Conditional and Object-Centric Gaussian Splatting World Models for Rigid Objects
}

\author{Jens U. Kreber,
Lukas Mack,
Joerg Stueckler%
\thanks{
This work has been supported by Hightech Agenda Bayern and Deutsche Forschungsgemeinschaft (DFG) project no. 466606396.
We gratefully acknowledge the HPC resources provided by the Erlangen National High Performance Computing Center (NHR@FAU) of the Friedrich-Alexander-Universität Erlangen-Nürnberg (FAU) under the BayernKI project v119ee. BayernKI funding is provided by Bavarian state authorities.
}
\thanks{
All authors are with the Intelligent Perception in Technical Systems Group, University of Augsburg, Germany {\tt\small firstname.lastname@uni-a.de}.
Project website: \url{https://embodiedvision.github.io/mro-gwm/}
}
}

\begin{document}

\bstctlcite{IEEEexample:BSTcontrol}

\maketitle
\thispagestyle{empty}
\pagestyle{empty}

\begin{abstract}
World models enable intelligent agents to predict the consequences of their actions on the environment.
In this paper, we propose Multi Rigid Object Gaussian World Model (MRO-GWM), a novel model that learns action-conditional dynamics of rigid objects in 3D.
Using object-centric Gaussians, we can represent arbitrary object shapes and multi-object scenes.
We develop a novel temporal-spatial transformer architecture that predicts future rigid body motion from a history of object Gaussians and future actions.
Objects are represented by Gaussians in an object-centric frame, which allows to describe object motion as a rigid body transformation.
Our model is trained on reconstructions from multiple viewpoints, which requires the model to also handle partial observations of objects due to occlusions.
We analyze prediction performance of our approach on synthetic datasets composed of typical household objects with multi-object dynamics and interactions by a robot end effector.
We also evaluate our model in model-predictive control for non-prehensile manipulation in simulation and for image-based tracking and prediction of object motion in simulated and real-world scenes.
\end{abstract}


\section{Introduction}

Intelligent agents that interact physically with the environment need world models to predict the consequences of potential actions in perceived scenes and to plan ahead.
World models capture task-agnostic properties of the environment and therefore form a keystone of generalization abilities.
A challenging problem for world modeling are rigid-body dynamics in robot manipulation scenes where many objects of highly varying shapes are closely placed next to each other.
An interaction with one object might have indirect effect on surrounding ones and the objects might not be fully observable due to occlusions.

Our novel Multi Rigid Object Gaussian World Model (MRO-GWM) represents the scene using object-centric 2D Gaussians~\cite{ObjectGS} and learns the dynamics on this representation end-to end.
We choose Gaussians as our shape representation, since they capture geometry for modeling object interactions and enable real-time rendering for object detection, pose estimation, and tracking.
We propose a novel temporal-spatial transformer architecture which predicts future object poses and is conditioned on a history of object and end-effector states.
The state history is obtained by transforming per-object Gaussian-based shape representations by rigid object and end-effector poses.
We also use the representation for object pose estimation and tracking with a render-and-compare approach.

We evaluate our method on simulated tabletop scenes containing closely placed household objects of varying shapes.
We compare model variants and ablations regarding prediction performance on unseen objects.
Furthermore, we evaluate model-predictive control with our model in simulation for two non-prehensile manipulation tasks and demonstrate that it can successfully solve the tasks in several multi-object scenes.
Finally, we demonstrate our model for image-based tracking and prediction of object motion in simulated and real-world scenes.

In summary, our contributions are as follows: 
(1) We propose to use object-centric Gaussian splatting as scene representation for seamlessly enabling dynamics prediction and scene perception.
(2) We introduce a novel transformer architecture for dynamics prediction that operates on this representation using temporal and spatial attention as well as a new temporal-spatial attention layer.
(3) We integrate and demonstrate our world model for image-based tracking and prediction of object motion and model-predictive control in non-prehensile manipulation.

\begin{figure}[tb]
    \centering
    \includegraphics[width=0.98\linewidth]{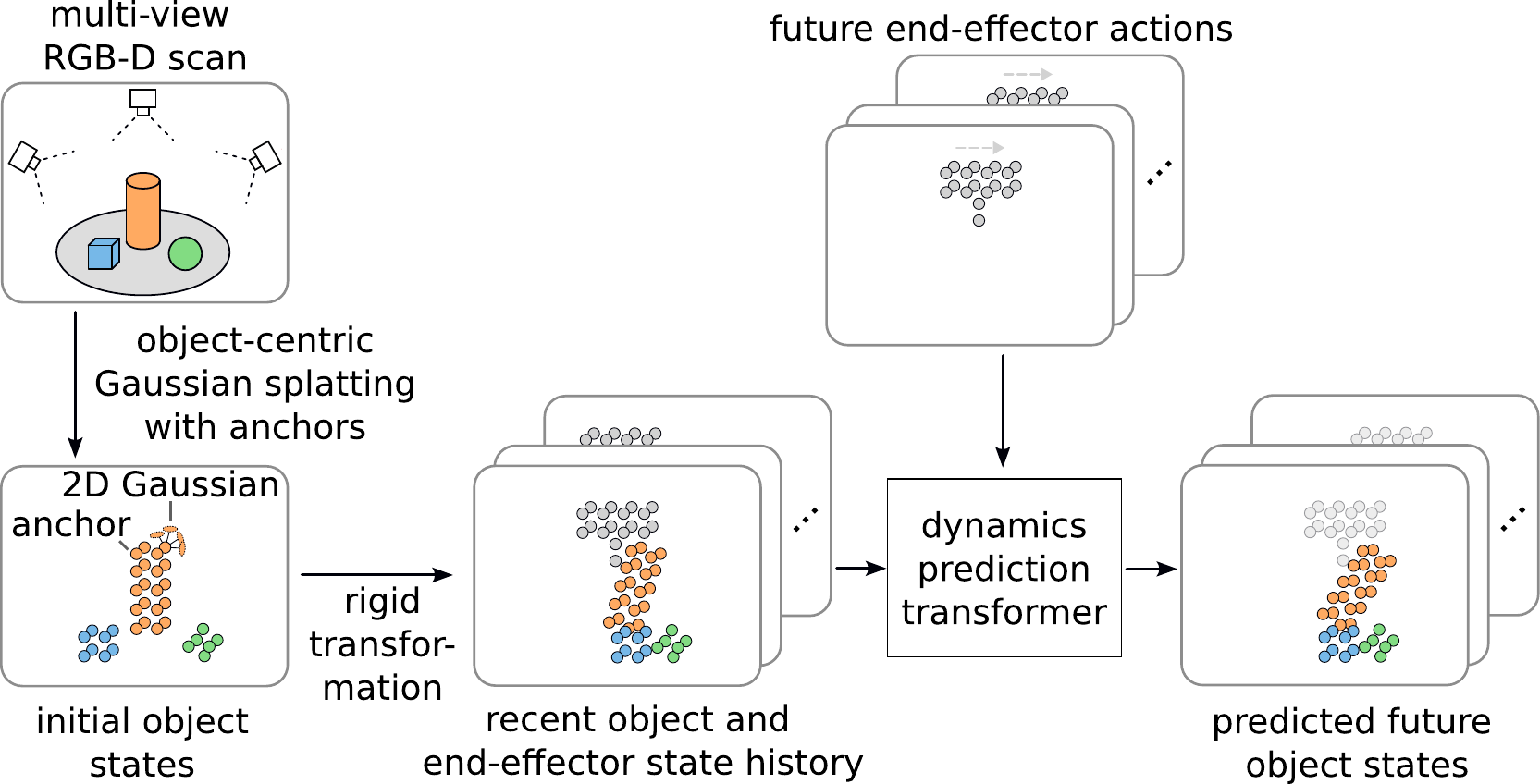}
    \caption{Method overview.
    Gaussian representations are obtained by object-centric Gaussian splatting.
    The Gaussians are efficiently encoded in anchors.
    The Gaussian-based shape representations are rigidly transformed by the history of recent object and end-effector poses as input to a transformer that predicts the future object poses. 
    Additionally, we condition the dynamics prediction on the future robot end-effector poses as actions by also inputting their Gaussian-based shape representation into the transformer.  }
    \label{fig:teaser}
\end{figure}

\section{Related Work}

Several approaches for learning world models have been proposed in recent years in the machine learning, robot learning, and computer vision communities.
One approach is to learn  action-conditional video prediction models~\cite{hafner2019_planet,bruce2024_igenie}.
The approach in~\cite{compnerfdyn} learns a 3D world model for multi-object interactions, but uses implicit NeRF representations of object shape.
Gaussian representations provide an alternative explicit 3D representation which has recently attracted attention. 
Gaussians are specifically suitable for real-time rendering from novel view points and therefore enable the detection and tracking of the 3D pose of objects.
Moreover, the Gaussians represent object geometry which can be used for learning physical dynamics.
In this work, we demonstrate both aspects by image-based tracking and prediction of object motion and using the learned model for model-predictive control.

\paragraph{Integrating Physics Models with Gaussians}
Some approaches integrate physics models with Gaussian representations to enable dynamics prediction.
PIN-WM~\cite{li2025_pinwm} use differentiable rendering of Gaussians and differentiable rigid-body physics~\cite{belbute-peres2018_difflcp} to infer physical parameters from visual observations.
ContactGaussian-WM~\cite{wang2026_contactgaussianwm} uses a complementary-free differentiable physics engine to propagate objects represented by 3D Gaussians.
Differently to our setting, these works only consider single-object scenes.
Rigid objects and deformables are simulated in a differentiable particle physics simulation by Embodied Gaussians~\cite{abou-chakra2025_embodiedgaussians}. 
It couples Gaussians for visual scene representation with particles of a point-based dynamics simulation.
Visual reconstruction errors are mapped to physical forces on the particles.
Related approaches, such as~\cite{zhang2024_physdreamer,xie2024_physgaussian,zhao2025_physsplat}, integrate physics simulation using the material point method~\cite{hu2018_mlsmpm} for deformables into a 3D Gaussian reconstruction of a scene.
Physics-based simulation is, however, restricted to the model class of the simulator and cannot cope with strong violations of the physical model.
Also, differently to our learning-based model, these approaches require that the scene and objects shapes are fully known or observed so that physics-based simulation becomes feasible.

\paragraph{Predicting Neural Physical Dynamics for Gaussians}
Several recent works learn deep learning-based dynamics prediction for Gaussian scene representations~\cite{lu2024_manigaussian,yu2025_manigaussianpp,pan2025_selfcorrecting,lu2025_gaussianworldmodel,chai2025_gaf,zhang2024_dyn3dtracking,tseng2025_gaussiansplattingvmpc}.
The models are demonstrated for imitation learning, model-based reinforcement learning, model-predictive control or failure detection when executing motion plans.
3DGSim~\cite{zhobro2025_3dgsim} trains transformers for simulating the dynamics and deformations of objects  represented as 3D Gaussians.
The transformer aggregates unstructured point clouds from multiple time steps into a single-step per-point prediction.
Our approach uses a spatial aggregation whose structure is shared over time and predicts multi-step per-object rigid pose changes.
Differently to the above approaches, our approach estimates an object-centric Gaussian scene representation and predicts rigid body motion of the objects using a transformer architecture.
Neural Gaussian Force Fields (NGFF~\cite{li2026_ngff}) learn an object-centric dynamics model for multi-object scenes.
They predict forces acting on objects with a neural interaction network architecture and use ODE solvers to determine object motion and deformations.
Contrary to our approach, they do not model robot actions by end-effector motions.
Furthermore, the approach fills occlusions in a feed-forward multi-view Gaussian reconstruction by shape completion.
We train a transformer model on partial shapes to implicitly model unobserved parts.

\section{Method}
\begin{figure}[tb]
    \centering
    \vspace{2mm}
    \includegraphics[width=0.98\linewidth]{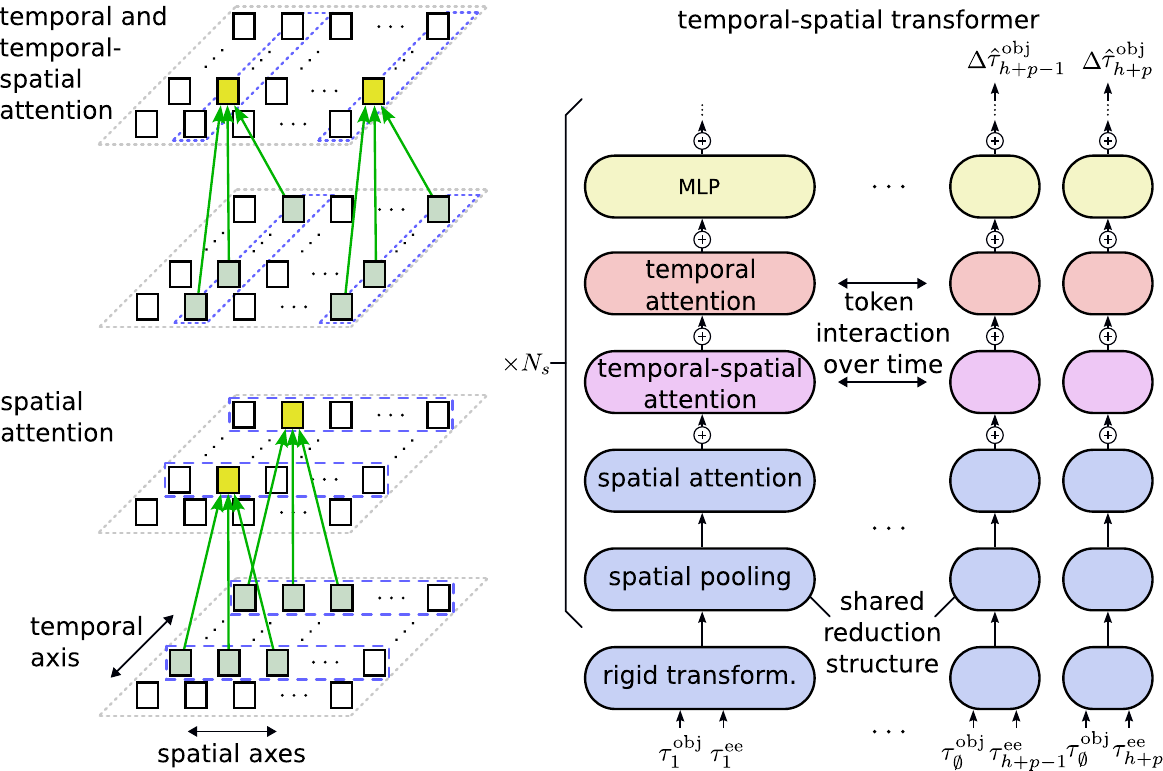}
    \caption{Dynamics prediction transformer. Top left: temporal and temporal-spatial attention relates all spatially corresponding tokens along the temporal dimensions.
    Bottom left: spatial attention relates tokens at the same time step which are in a local spatial neighborhood.
    Right: temporal-spatial transformer:
    Spatial grid pooling and spatial attention blocks are combined with temporal attention and a newly proposed temporal-spatial attention layer.
    Given a history and future end-effector poses, future object poses are predicted. $\hat{\tau}^\text{obj}_i$ indicate predicted poses and $\tau^\text{obj}_\emptyset$ a fixed dummy pose.  }
    \label{fig:architecture}
\end{figure}

We propose a novel approach for world modeling by predicting rigid-object dynamics in multi-object scenes conditioned on robot actions based on a Gaussian shape representation.
Our world model represents the geometry and appearance of objects as Gaussians and predicts action-conditional object dynamics using a transformer-based deep learning architecture.
By using Gaussians, we can render objects from novel view points for object perception.
Moreover, this shape representation provides object geometry information for dynamics and interaction modeling.
The representation can be reconstructed from multi-view RGB-D images.
A specific challenge when using Gaussian-based representations is to devise an efficient and suitable deep learning architecture.
To this end, we propose a novel object-centric, transformer-based architecture that operates
on a compressed anchor representation in a window of time steps using spatial and temporal attention layers.
It predicts future object pose trajectories given robot end-effector actions.
By using an object-centric representation, we can transform the shape representation of an object by its observed or predicted $SE(3)$ pose and predict relative motion in $SE(3)$.

\subsection{Scene Representation By Object-Aware Gaussian Splatting}

We follow ObjectGS~\cite{ObjectGS} to extract the Gaussian representation of the objects in the scene from multi-view image by Gaussian splatting~\cite{3dgs}.
We assume access to multiple posed RGB-D images~$I \subset \mathbb{R}^{n_\text{img} \times w_\text{img} \times h_\text{img} \times 4}$ with corresponding camera poses~$\xi^\text{cam}_{i} \in SE(3)$ in the world frame.
We also assume corresponding object segmentation masks $O \subset \mathbb{N}^{n_\text{img} \times w_\text{img} \times h_\text{img} \times N_\text{obj}}$ to be known, e.g., we use ground-truth or SAM3~\cite{sam3} object masks in our simulation or real-world experiments, respectively.
ObjectGS is an extension of ScaffoldGS~\cite{scaffoldgs}, which does not optimize 3D Gaussians directly, but represents the scene by a grid of voxels and representative points in each voxel called anchors.
Each anchor carries a latent feature which encodes a fixed number of Gaussians in a local neighborhood of the anchor.
The location of the Gaussians is represented by offsets to the anchor position and scale factors for the offsets.
The Gaussians are decoded with an MLP and the anchor features are optimized with a splatting rendering loss.
ScaffoldGS demonstrates that this provides a compressed representation, while maintaining similar rendering quality than direct Gaussian splatting.
ObjectGS extends ScaffoldGS by assigning object identities (ids) to each anchor and corresponding decoded Gaussian based on object segmentation masks for each image in a voting process.
Since we consider scenes which are closely populated with objects, we only consider an object id vote, if the projected depth of the anchor is within a \SI{1}{\centi\meter} range of the measured depth at the pixel.
We use the 2D Gaussian variant~\cite{2dgs} of ObjectGS, which we observed to represent surfaces better than 3D Gaussians~\cite{ObjectGS} in our setting.

\paragraph{Scene-Agnostic Anchor Features}
We aim at learning dynamics prediction which generalizes across various objects and scenes.
However, ObjectGS trains encoders for the anchor features for each scene separately.
We generalize the method to various objects and scenes by training shared feature encoders across multiple scenes.
We split the anchor features into 16 dimensions for spatial information, i.e. scale, rotation, opacity, and 16 for color.
MLPs are shared for the former and distinct per scene for the latter.
Dynamics prediction only uses the spatial part.

\paragraph{Robot End-Effector Representation}
Additionally, we render multiple views of a robot end effector and reconstruct its Gaussian representation in a similar way like the scene.
This allows us to place the end effector spatially into the scene and vary its pose.
Note that we use a distinct MLP to model the end effector with higher quality. 
We did not observe a clear loss in prediction performance compared to using the shared MLP of the scenes.

\subsection{Dynamics Prediction}

We implement dynamics prediction in our world model by a transformer-based architecture.
It predicts future motion of the objects conditioned on the future motion of the robot end-effector.
To this end, we input the Gaussian shape representations in a history of object~$\tau^\text{obj}_{1:h}:=\tau^\text{obj}_{1}, \ldots, \tau^\text{obj}_{h}, \tau^\text{obj}_{i} \in SE(3)$ and end-effector poses~$\tau^\text{ee}_{1:h}:=\tau^\text{ee}_{1}, \ldots, \tau^\text{ee}_{h}, \tau^\text{ee}_{i} \in SE(3)$ to the transformer at a certain temporal model rate, where~$h$ is the length of the history.
Given the end-effector Gaussian shape representation in future poses, the transformer predicts the future relative sequential $SE(3)$ motion of the objects~$\Delta \tau^\text{obj}_{h+1:h+p}$ within a prediction horizon~$p$ at the same rate as the input history.
We represent the object shape in their object-centric coordinate frames, but model object dynamics in the world frame.
Notably, the object frame can be freely chosen during the initial scene reconstruction, i.e., no canonical object frame is required, since only relative motion is predicted.

\paragraph{Transformer Inputs}
Central to our architecture is the idea to rigidly transform the shape representations of each object according to their observed poses in the history and input these as tokens into the transformer.
By using relative operations on 3D positions in the transformer, the transformer becomes translation-invariant.
To input actions, we also transform the shape representation of the end effector with its poses.
The $SE(3)$ motion of the objects in future time steps is predicted by the transformer.
To input their initial tokens to the transformer, we use the identity transform.
This results in $h+p$ scene representations for different object and end-effector poses.
We support Gaussians and anchors as input tokens to the transformer.
For the anchors, we can just provide their position or also their Gaussian attributes (i.e., offsets, scale, latent features).
In all cases, we also embed the corresponding object index and a Boolean representing if the token is part of a history or future time step and concatenate these embeddings.
The scene representation can have a large spatial extent due to the background being included.
However, it is reasonable to assume that interactions can only occur close to objects.
In order to safe unnecessary computation, we do not input background anchors that are further than some threshold (e.g., \SI{10}{\centi\meter} in our experiments) away from object or end-effector anchors.

\paragraph{Temporal-Spatial Transformer Architecture}
We choose transformers for our problem of dynamics predictions, since they relate features within and between objects and time steps which is important to model the object motion and their physical interactions.
The dynamics prediction by the transformer should be translation-invariant in the choice of origin of the world frame.
Moreover, the transformer needs to operate on unordered sets of spatial tokens with arbitrary continuous positions.
To tackle these challenges, we use the spatial attention mechanism of PTv2~\cite{ptv2} for processing the input tokens.
In this spatial attention, neighboring points exchange information to form new output features.
For each point, the attention is computed over its k-nearest neighbors, where the position is only processed in spatial relation features, rendering the representation translation-invariant.

For increasing the efficiency of our model, we support an initial per-object encoding and grid pooling of the inputs to a certain intermediate input resolution using PTv2.
Next, our model uses~$N_s$ stages of several attention layer operations in the spatial and temporal dimensions to form features of the motion and interaction of tokens between objects and predict the future relative motion of the objects.
The token features in a stage are processed across all time steps simultaneously.
Each stage has the form~$\operatorname{s}_i = \operatorname{b}_{i,N_b} \,\circ\, \ldots \,\circ\, \operatorname{b}_{i,0} \,\circ\, \operatorname{pe} \,\circ\, \operatorname{pool}_i$, where $\operatorname{b}_{i,j}$ are individual blocks and $N_b$ is the number of attention blocks in a stage.
The stage begins with downsampling the input tokens using a pooling layer~$\operatorname{pool}_i$.
By this, the stages spatially aggregate the Gaussian-based representation of the objects successively into a single abstract token feature per object from which the relative object motion is predicted.  
Similar to pooling in PTv2, it combines features from points in the same grid cell and averages the respective coordinates per object.
The pooling is performed in local object coordinates at the resolution of each stage.
Besides saving computation, this also establishes correspondence between the tokens of different time steps, which simplifies the attention calculation along the temporal dimension.
After the pooling, a positional encoding~$\operatorname{pe}$ for the time index is added.
We do not add a positional encoding of the token coordinates in the world frame to keep the architecture translation-invariant.

Subsequently multiple residual blocks~$\operatorname{b}_{i,j}$ with several residual layers are applied before proceeding to the next stage.
For determining features of the temporal evolution of the scene, we relate token features across the temporal dimension in a temporal attention layer using standard self-attention~\cite{transformer} over time.
The coordinates of the tokens are not used in this attention.
For modeling object interactions and aggregating information spatially across tokens between objects in each time step, we use a spatial attention layer based on PTv2's spatial attention mechanism.
Lastly, we also determine features of the temporal evolution of the tokens which takes spatial motion of the tokens into account.
To this end, we propose a novel temporal-spatial attention layer, in which each token in a time step is related to its corresponding tokens at different time steps.
We use PTv2's spatial attention mechanism based on the token coordinates.

We stack the layers in the following order: spatial attention $\rightarrow$ temporal-spatial attention $\rightarrow$ temporal attention $\rightarrow$ MLP,
where the attention operations are followed by a pointwise MLP operation.
For the last stage, we choose a grid cell size that spans the whole scene to aggregate a single token feature for relative motion prediction.
The motion is predicted using a linear layer.
Note that we discard the predictions for background and end effector as we assume their motion to be fixed or given as actions, respectively.

\paragraph{Prediction Representation and Losses}
We train our model to predict the relative motion in position and orientation in world coordinates between two time steps across the prediction horizon for each object.
Following~\cite{hitchhikerrotations}, we represent rotations as 9D rotation matrices and apply orthogonalization on the predictions to achieve valid 3D rotations.
We use the squared Frobenius norm of predicted and ground truth matrices, which corresponds to the commonly used squared chordal distance in the case of rotation matrices~\cite{hitchhikerrotations}.
Since we predict relative rotation, which is often small, we need a way to normalize this data to train the network.
We apply a vector-like normalization, where we first subtract the identity from the target relative rotation matrix $\mathbf{R} \in SO(3)$ and then divide each component by its standard deviation over the training set.
To obtain a valid rotation from a prediction, the normalization process is reversed and the resulting matrix orthogonalized.
Note that we determine the loss in the normalized space, before orthogonalization.
We experimented with other loss types such as a geodesic loss and other loss schemes, such as computing the loss on orthogonalized rotation matrices or first orthogonalizing and normalizing again.
However, the described method performed best in our experiments.
For the relative position, we divide the ground truth motion by its mean norm over the training set and compute the squared l2 norm against the prediction.

\subsection{Object Pose Estimation and Tracking}
We detect and track the 3D pose of objects by rendering the Gaussians in a render-and-compare approach.
We augment FoundationPose~\cite{wen2024_foundationpose} to use the renderings of the objects.

\section{Experiments}

\subsection{Experiment Setup} 
\label{sec:environment}

\paragraph{Scene generation and datasets}
We train and evaluate our model on datasets obtained using the Maniskill simulator \cite{maniskill}.
We consider tabletop scenes populated with varying counts of 1 to 5 of objects from the YCB object set~\cite{calli2015_ycbobjects}.
We generate scenes where objects are closely positioned next to each other and therefore likely interacting during manipulation by the robot.
We use disjoint object pools for our train, validation (val) and test datasets to analyze generalization to unseen objects.
The YCB object set contains 75 objects from which we use 65 for training, 5 for validation, and 5 for testing.
To asses the importance of object variety seen during training, for one ablation we only use 16 of the 21 objects from \cite{xiang2018_posecnn}.
The remaining 5 objects are used for our val set.
For the test pools, we sample one object from each of the categories proposed by~\cite{calli2015_ycbobjects}.
We augment our train and val sets by generating additional scenes that use the exact same objects as the initially sampled ones, but new random pose sampling and interaction data collection.
For these augmented scenes, we use the same splatting representation as the initial scene, since only object poses are changed, thereby obtaining more interaction data with the same number of splatted scenes.
Our main training set contains 100 original scenes and for each of those 100 augmented ones.
To avoid sampling scenes with heavily occluded objects which can barely be reconstructed, we reject scenes where objects are not visible by at least 50\% and 1024 pixels in at least 25\% of the views.
We model a Franka Panda robot with a stick-like end-effector, similarly to the tasks already shipped with Maniskill.
The end-effector actions are relative 3D position changes executed by an inverse kinematics controller.
Interaction data is generated by sampling points in the vicinity of the objects and moving the end effector to these points with a constant speed of~$5$\,cm/s.
By this, the robot moves around or pushes objects.
The simulator is stepped with $\SI{20}{\hertz}$.

For prediction evaluation, we create a test set with 20 scenes for each object count from 1 to 5, resulting in 100 scenes total.
We do not use augmentations, but sample 3 different interaction trajectories.
For planning evaluation, we create 8 scenes per object count from 1 to 5, resulting in 40 in total.
To capture a scene, we obtain 32 images at two different elevation angles, constant radius and equally spaced azimuth angles.
We perform object-aware Gaussian splatting as described above with a default anchor resolution of $\SI{1}{\centi\meter}$ for 5,000 steps.
We use rendered masks and depths as input.

\paragraph{Model training}
As training examples, we split the recorded interaction trajectories into small chunks according to the model's prediction horizon (default $p=4$) and modeling frequency (default \SI{5}{\hertz}).
We use a fixed history size of $h=3$.
We find hyperparameters of our model such as number of stages, pooling size and number of channels by an automated hyperparameter search with limited training budget using the two objectives prediction error and training time.
We select two variants from the Pareto front: Our base model (B), which we use as default for all experiments, with best error using 3 stages and no object encoder as well as a fast model (F) with fewer parameters, 2 stages and a 1-stage object encoder.
We conduct training with a batch size of 45 for 60,000 steps, which takes approximately \SI{10.7}{\hour} for the base and \SI{2.7}{\hour} for the fast model on an Nvidia H100.
By default, we input anchors with Gaussian attributes.

\paragraph{Prediction performance evaluation}
We assess prediction performance on trajectory chunks of varying sizes which requires different counts of recurrent model calls.
We compare the predicted poses at certain prediction horizons with the ground truth and report the Euclidean distance between positions and the geodesic angle between rotations.
We provide a study on variants and ablations of our model to assess design choices.
Note that there is no other work available that allows for direct comparison in our setting of multi-object dynamics with partial observations.
For~\cite{compnerfdyn}, which is closest in setting, there is no code available.
It is also not feasible to compare with~\cite{zhobro2025_3dgsim}, since code and rigid object datasets are not available. Furthermore, the approach is evaluated on single-object scenes and reports visual fidelity instead of trajectory accuracy. 
The multi-object dynamics learning approach in~\cite{li2026_ngff} does not directly support conditioning on end-effector position actions.

\paragraph{Planning performance evaluation}

We evaluate planning performance with our learned world model in two tasks using a variant of iCEM~\cite{pinneri2021_icem}.
It performs sampling-based optimization of keypoint-parametrized end-effector trajectories with respect to a set of task-specific costs over a planning horizon.
The end-effector keypoints are linearly interpolated at a constant velocity for execution and passing them into the dynamics prediction.
In task 1, the robot needs to push objects to goal locations which we consider successful if the distance is below \SI{1}{\centi\meter}.
We also provide the area under the curve (AUC) of the success rate for thresholds ranging from \SI{1}{\centi\meter} to \SI{5}{\centi\meter} in \SI{5}{\milli\meter} steps.
As objective function for planning, we minimize a combination of object-goal and end-effector-object distances.
In task 2, the robot needs to move all objects outside a center region on the table in the sidewards interval $[-\SI{15}{\centi\meter}, \SI{15}{\centi\meter}]$.
We evaluate the smaller distance from object center to either bound and sum these up for all objects as goal distance.
Success is determined on \SI{1}{\centi\meter} distance to bounds for all objects.
We also evaluate a success AUC metric for thresholds from \SI{0}{\centi\meter} to \SI{5}{\centi\meter} in \SI{5}{\milli\meter} steps.
As objective function for planning, we minimize a combination of the distances of the objects to the goal area, distances of the end effector to non-cleared objects, and a penalty for pushing objects far to the front or back of the table.
We also compare with a heuristic that selects the object with furthest goal distance and moves into it until reaching the goal area.
We run episodes for both tasks for 1200 environment steps, i.e., \SI{1}{\minute}.
Each task is evaluated over the 40 planning test scenes with 5 episodes per scene.

\paragraph{Tracking performance evaluation}
Tracking performance with our model is evaluated with a subset of 5 YCB objects without continuous rotational symmetries.
The rest is used for training the dynamics prediction from which we use 5 for tuning FoundationPose.
For each interaction trajectory, we perform pose estimation at the first step and subsequently track the objects.
We only consider trajectories up to the point where one object fails our visibility criterion introduced earlier.
We select a constant view point out of several candidates on a separate validation set with the highest fraction of visible object frames.
For tracking, we use a Gaussian splatting representation with voxel resolution 0.005 and no view dependence, which we found to perform best on the validation set.
We only evaluate tracking of the objects but not the end effector.

\subsection{Prediction Results}

\begin{figure}[t]
    \centering
    \vspace{2mm}
    \includegraphics[width=0.98\linewidth]{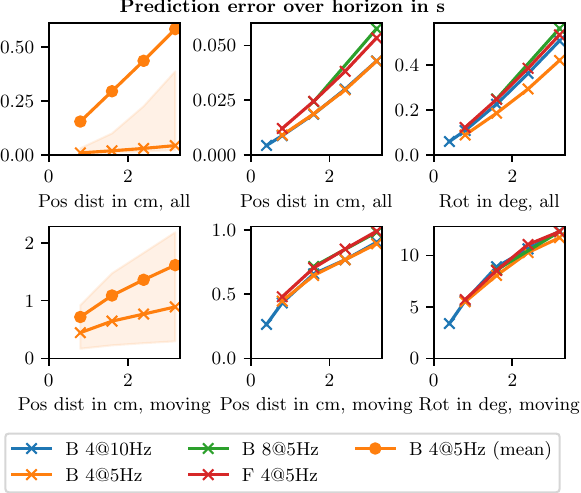}%
    \caption{Prediction errors over horizon for different model variants, considering either all poses at the prediction horizon (top) or only those where a ground truth movement for the respective object occurs during the horizon (bottom).
    The first column shows mean and median as well as the area between the 25\% and 75\% quantile.
    Other plots show only the median.
    Predictions of 5 model training seeds were used.
    The base model at $\SI{5}{\hertz}$ performs best.
    }
    \label{fig:pred-over-horizon}
\end{figure}
In \Cref{fig:pred-over-horizon}, we analyze the prediction performance in position and rotation error and of model variants with different time resolution and prediction horizons.
We distinguish results considering all errors at the prediction horizon (all) and those where a ground truth movement (pos. change $> \SI{1e-4}{\meter}$, rot. change $> \SI{1e-2}{\radian}$ for single step) happens during the horizon for the considered object (moving).
The spread between 25\% and 75\% quantiles can be explained by the variation caused by static and moving objects.
We also observe that the error is more pronounced in rotation which we attribute to object dynamics such as rolling or toppling over.
Regarding variations in model frequency and prediction horizon, we observe that for position error,~$p=4$ works equally well at $\SI{5}{\hertz}$ and $\SI{10}{\hertz}$, while~$p=8$ or the fast model (F) lead to a similar performance drop.
The base model (B) at $\SI{5}{\hertz}$ performs best in rotational error.

\paragraph{Model variants and ablations}
\begin{table}[tb]
    \vspace{2mm}
    \centering
    \footnotesize
    \setlength{\tabcolsep}{6pt}
    \begin{tabular}{lccc}
        \toprule
Variant & \makecell{Mean pos\\err in cm} & \makecell{Mean rot\\err in deg} & \makecell{Full std\\ pos, rot} \\
        \midrule
Base model & 0.717 (0.002) & 12.484 (0.099) & 0.86 \ 18.16 \\
Fast model & 0.758 (0.004) & 13.093 (0.082) & 0.89 \ 19.16 \\
Few objects & 0.816 (0.010) & 13.611 (0.140)& 0.94 \  19.34 \\
Anchor res 0.005 & 0.713 (0.004) & 12.571 (0.119) & 0.86 \ 18.38 \\
Anchor res 0.02  & 0.734 (0.002) & 12.473 (0.059) & 0.87 \ 17.99 \\
Explicit Gaussians & 0.719 (0.008) & 12.486 (0.168) & 0.87 \ 18.08 \\
No anchor attributes & 0.718 (0.003) & 12.468 (0.050) & 0.87 \ 18.06 \\
No attr., res 0.02 & 0.745 (0.006) & 12.693 (0.088) & 0.87 \ 18.44 \\
No temp.-sp. attn. & 0.754 (0.008) & 12.787 (0.107)& 0.87 \ 18.27 \\
Single obj. token & 0.827 (0.003) & 14.107 (0.069) & 0.92 \ 19.49 \\
Limited view & 0.724 (0.003) & 12.970 (0.106) & 0.88 \ 18.63\\
\bottomrule
    \end{tabular}
    \caption{Prediction performance of model variants and ablations, computed over moving chunks only.
    The mean over 5 model training seeds is shown with standard deviation over seeds in parentheses.
    The last two columns show the standard deviation over all values.
    We observe strong prediction performance drops for architectural ablations, the fast model variant and few training objects.
    Anchor attributes become relevant at coarser resolution.
    }
    \label{tab:pred-ablations}
\end{table}
In \Cref{tab:pred-ablations}, we compare our base model with several ablations.
The fast model provides less accurate prediction at lower inference time of \SI{0.063}{\second} at batch size 45 compared to \SI{0.168}{\second} for the base model.
The performance drop for few training objects demonstrates that a large object variety is important for generalization.
Performance is marginally better in position for smaller anchor size, but drops slightly for a resolution of 0.02, which is expected due to the coarser scene geometry representation.
Using Gaussians leads to similar prediction accuracy like the base model with coarser anchors, but requires \SI{0.326}{\second} inference time.
We observe that using anchor attributes brings an improvement for resolution 0.02, while it has similar performance to using no anchor attributes at resolution 0.01.
Removing the temporal-spatial attention layer leads to a strong performance drop.
Using only a single token per object strongly degrades accuracy, which is expected due to the complex geometry of objects in our datasets.
Limiting azimuth angles during reconstruction between 0 and 180$\unit{\degree}$ performs only slightly worse than training on the full set of views.
This indicates that the model to some extent learns to cope with partial shape information.

\paragraph{Qualitative results}
\begin{figure}[t]
    \centering
    \vspace{2mm}
    \setlength{\tabcolsep}{0pt}
    \renewcommand{\arraystretch}{0.}
    \begin{minipage}{\linewidth}
        \begin{tabular}{cccccc}
        \includegraphics[width=0.2\linewidth]{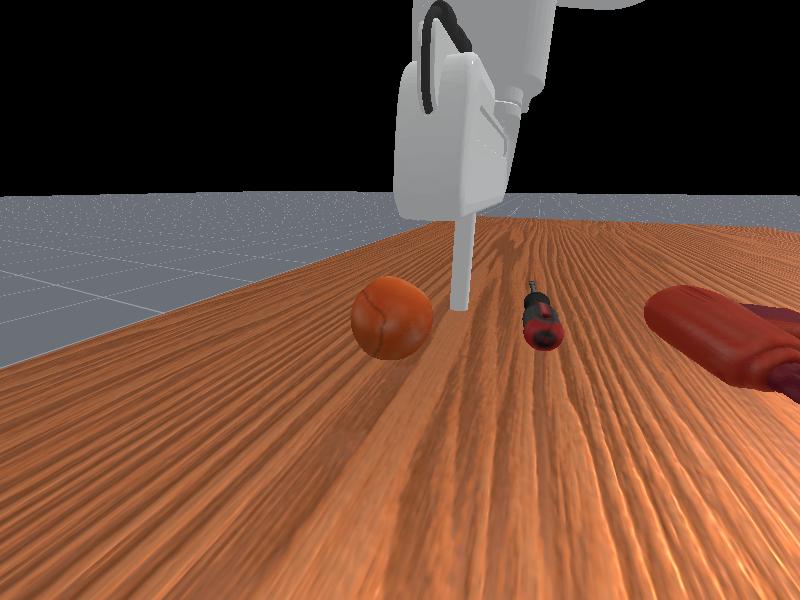} &
        \includegraphics[width=0.2\linewidth]{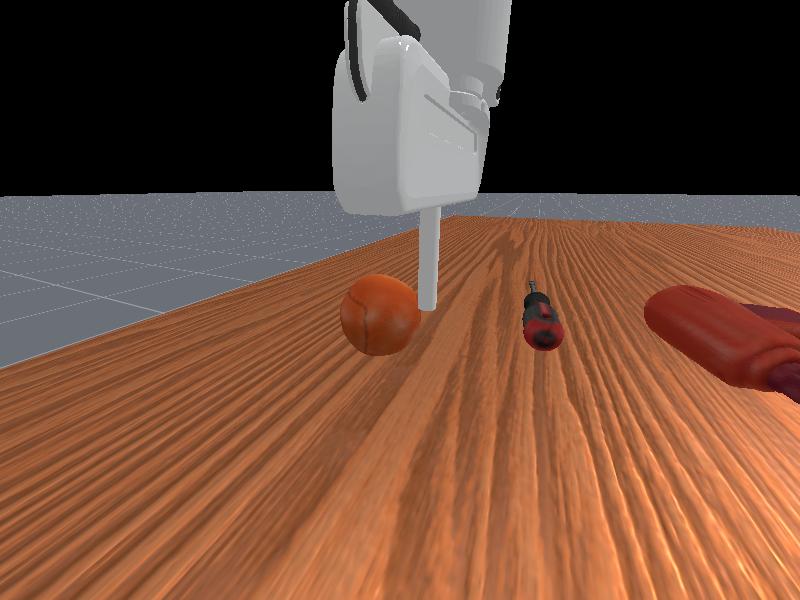} &
        \includegraphics[width=0.2\linewidth]{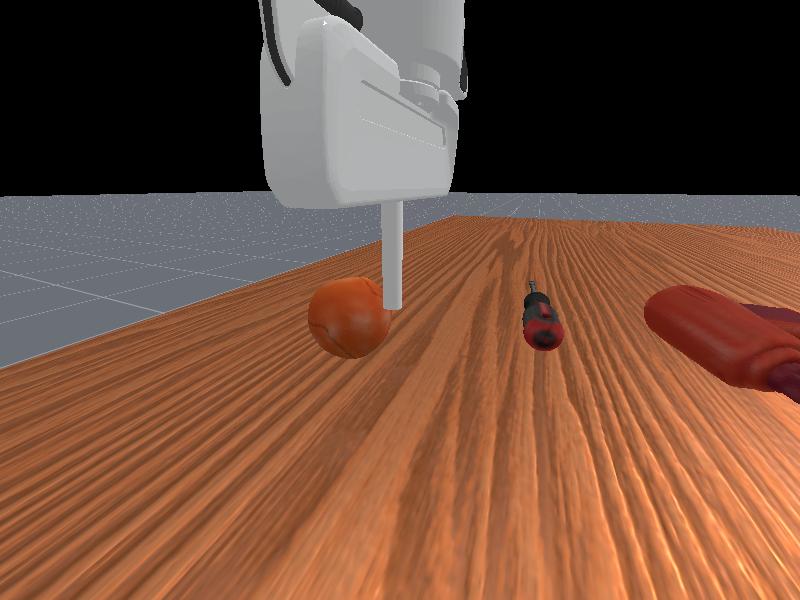} &
        \includegraphics[width=0.2\linewidth]{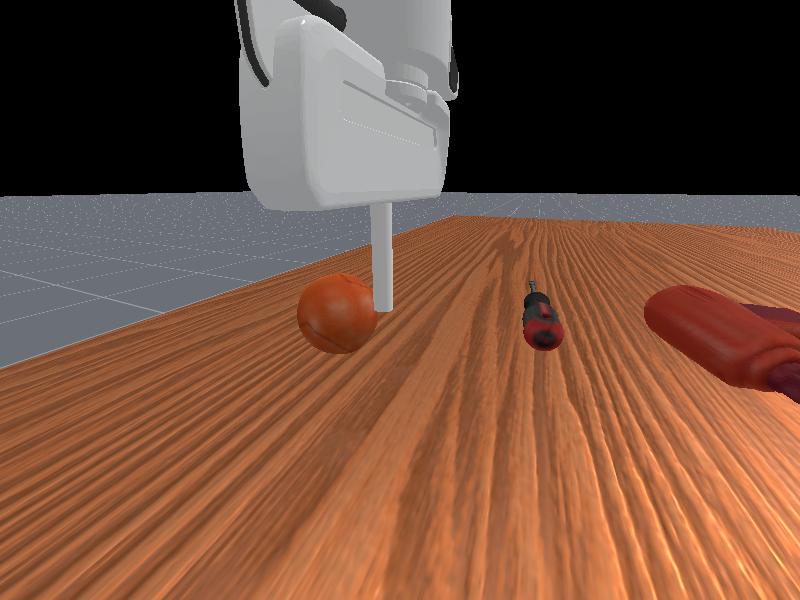} &
        \includegraphics[width=0.2\linewidth]{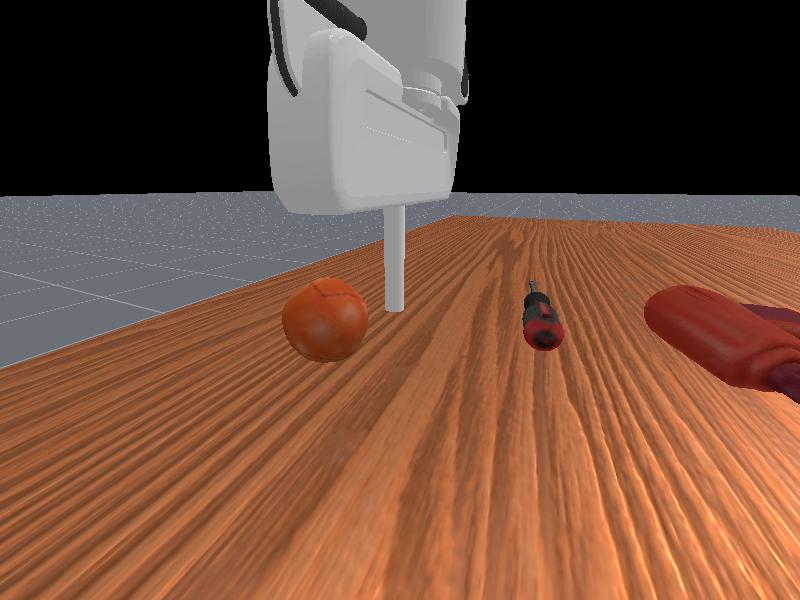}\\
        \includegraphics[width=0.2\linewidth]{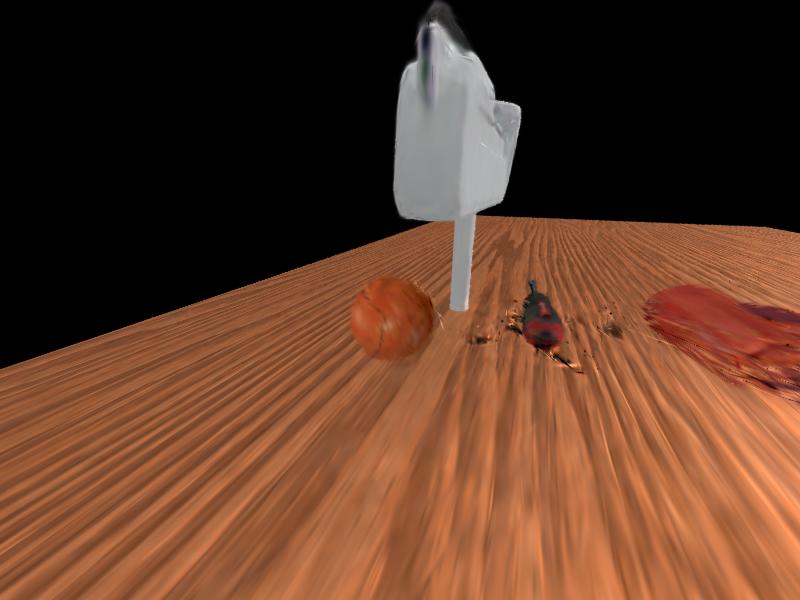} &
        \includegraphics[width=0.2\linewidth]{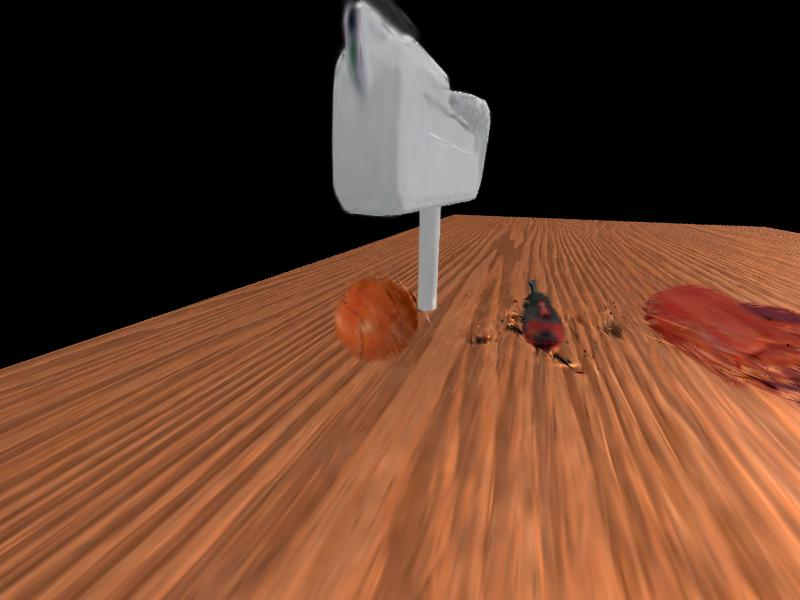} &
        \includegraphics[width=0.2\linewidth]{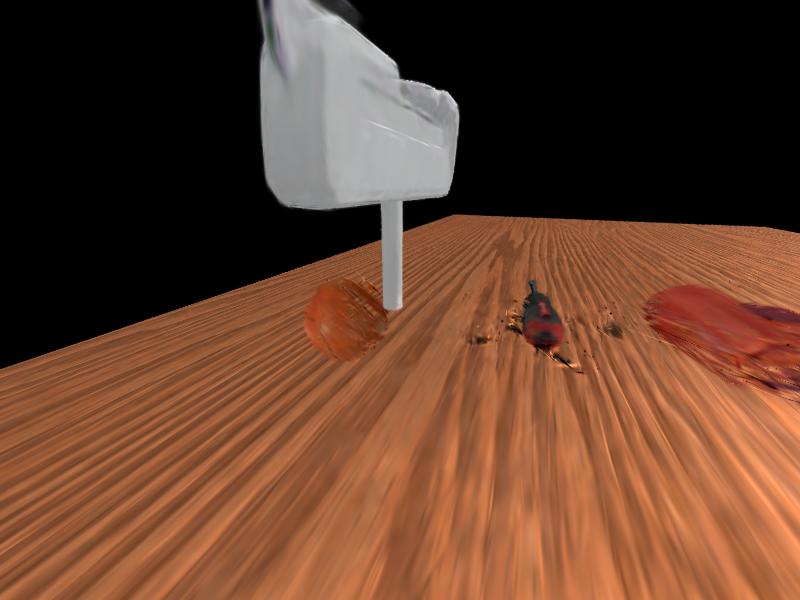} &
        \includegraphics[width=0.2\linewidth]{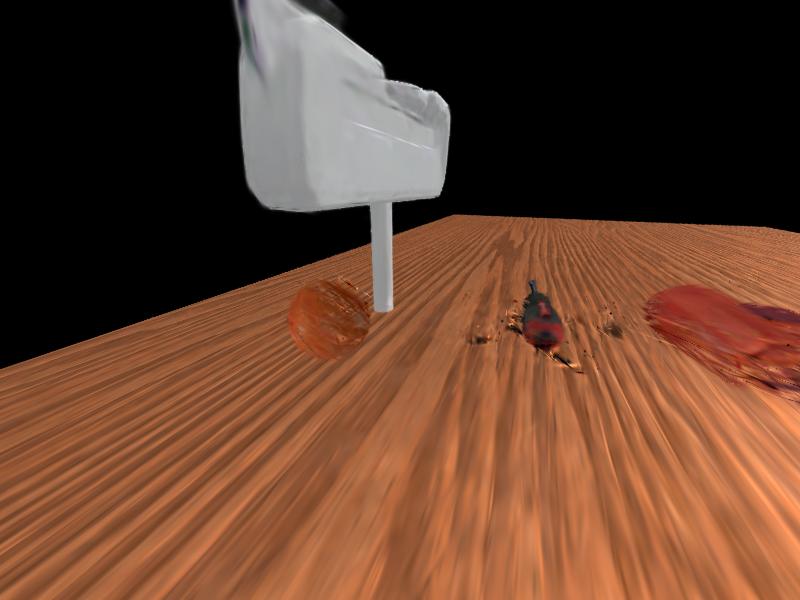} &
        \includegraphics[width=0.2\linewidth]{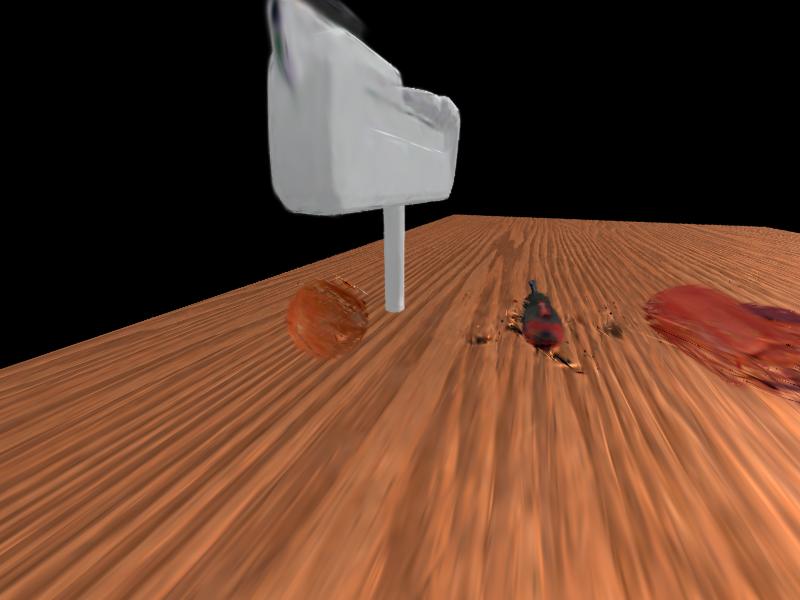}\\
        \end{tabular}
     \end{minipage}\hfill
     \vspace{0.1ex}
    \begin{minipage}{\linewidth}
        \centering
        \begin{tabular}{cccccc}
        \includegraphics[width=0.2\linewidth]{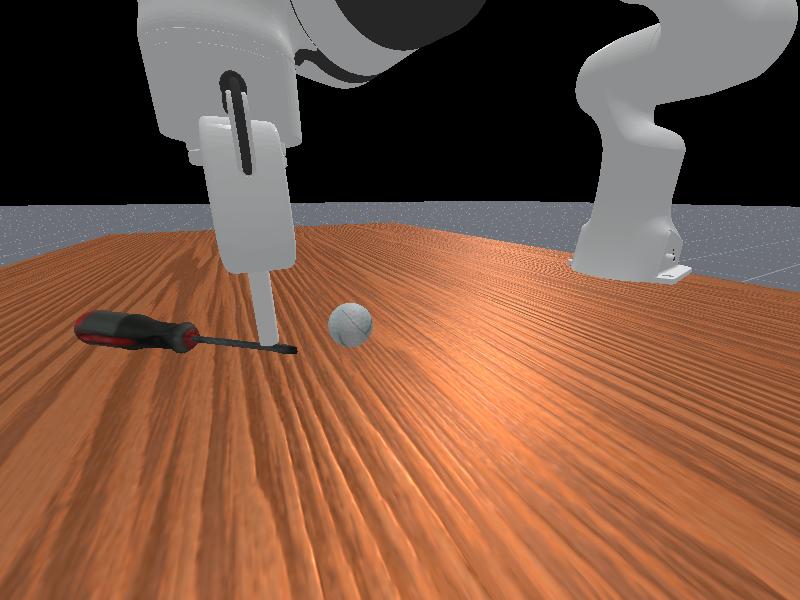} &
        \includegraphics[width=0.2\linewidth]{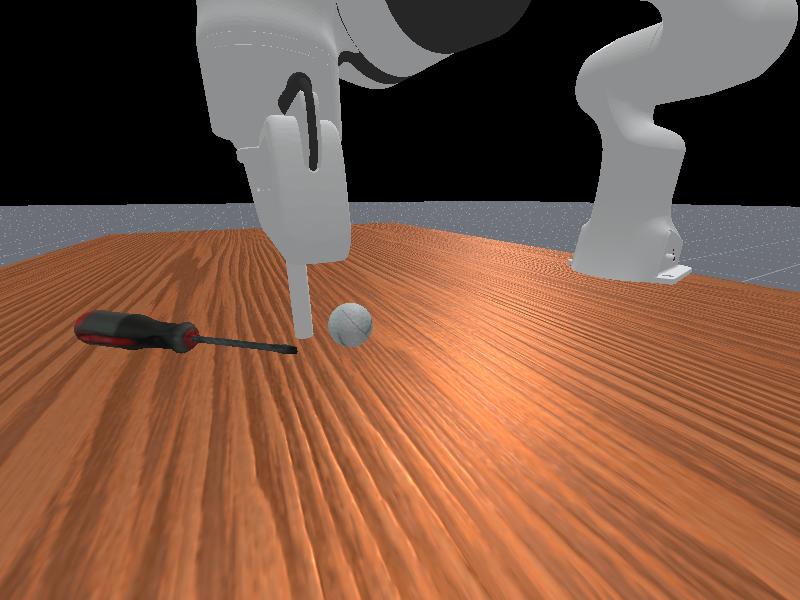} &
        \includegraphics[width=0.2\linewidth]{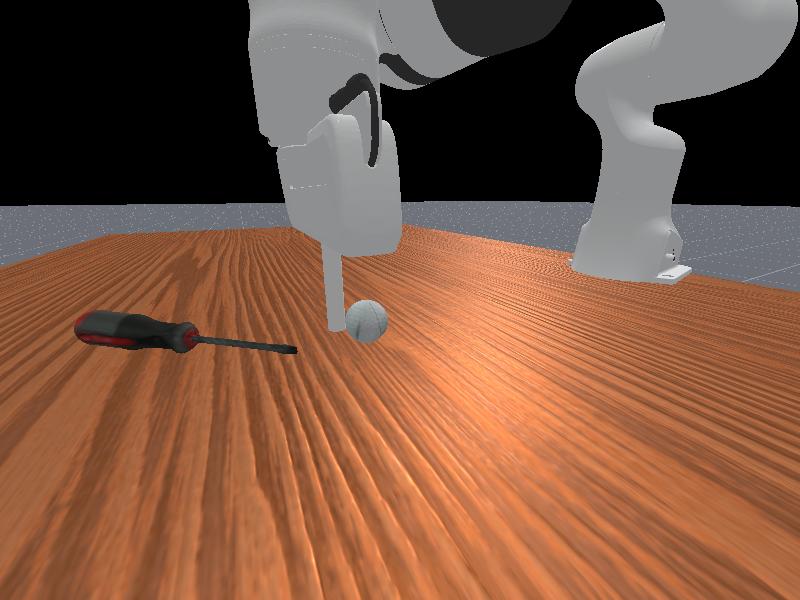} &
        \includegraphics[width=0.2\linewidth]{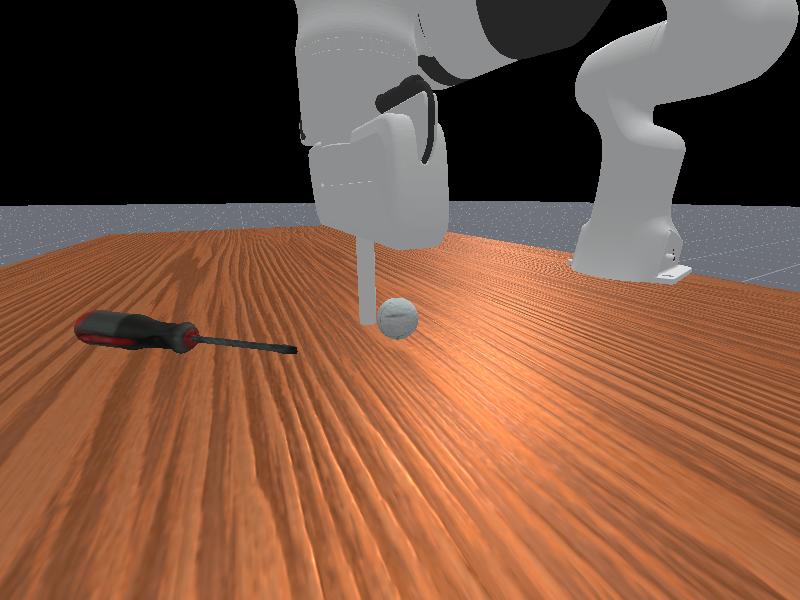} &
        \includegraphics[width=0.2\linewidth]{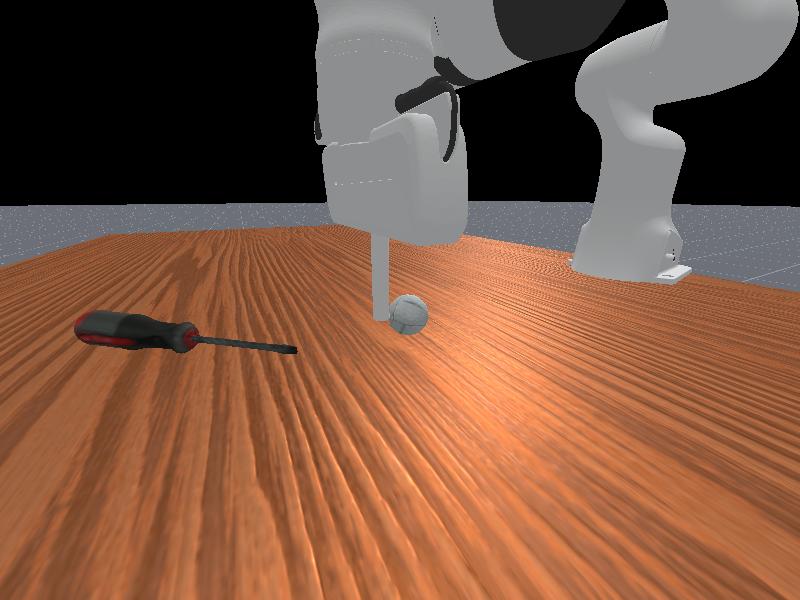}\\
        \includegraphics[width=0.2\linewidth]{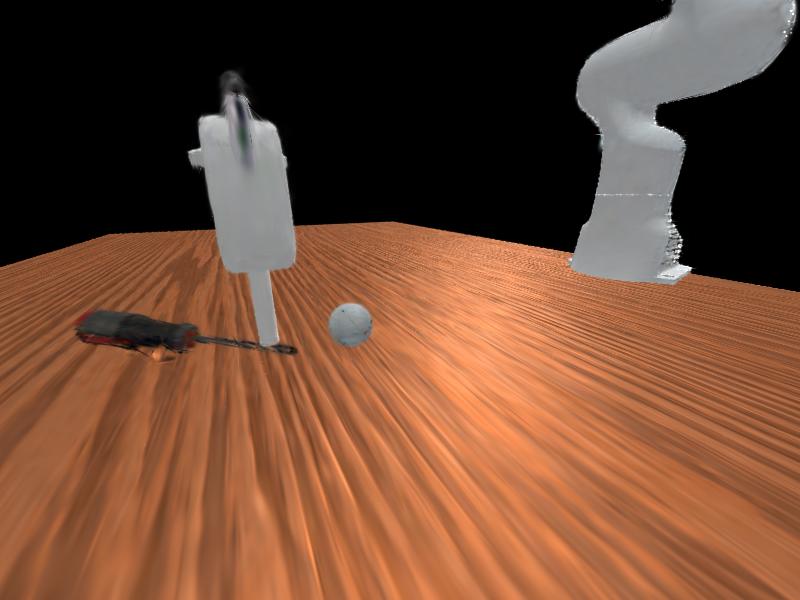} &
        \includegraphics[width=0.2\linewidth]{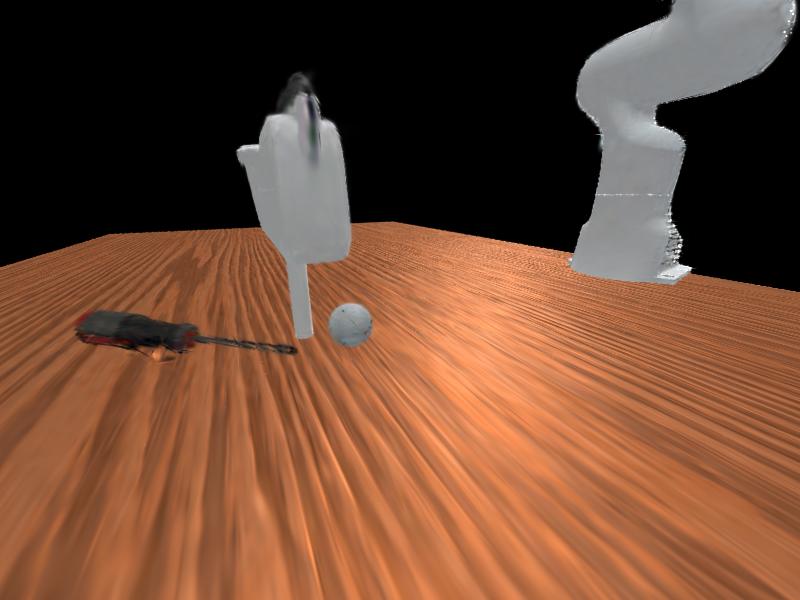} &
        \includegraphics[width=0.2\linewidth]{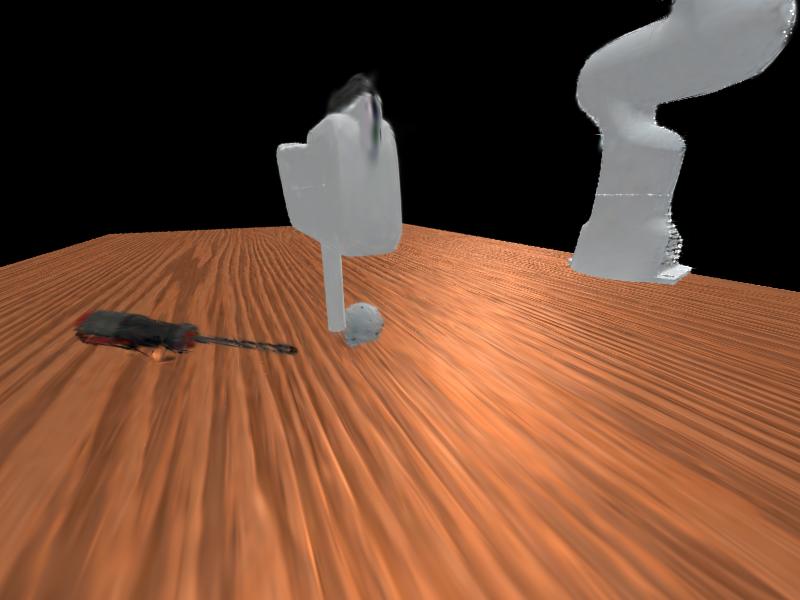} &
        \includegraphics[width=0.2\linewidth]{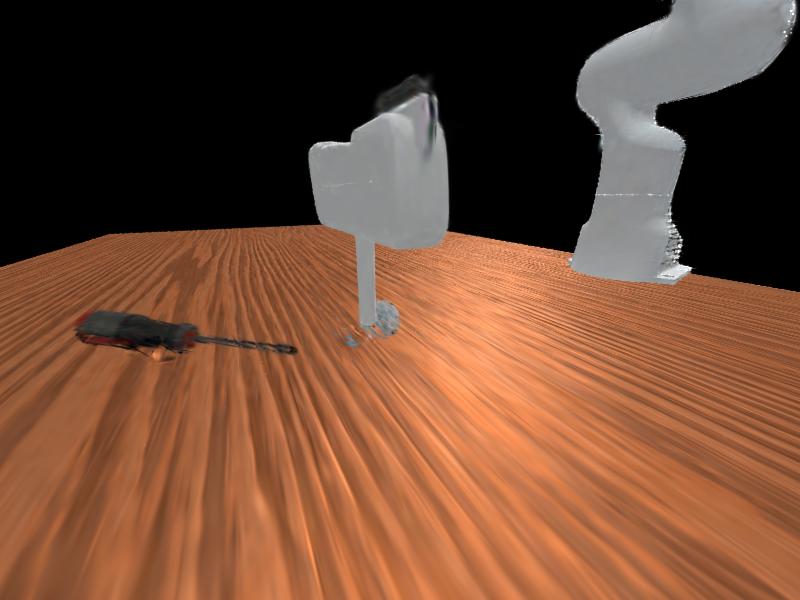} &
        \includegraphics[width=0.2\linewidth]{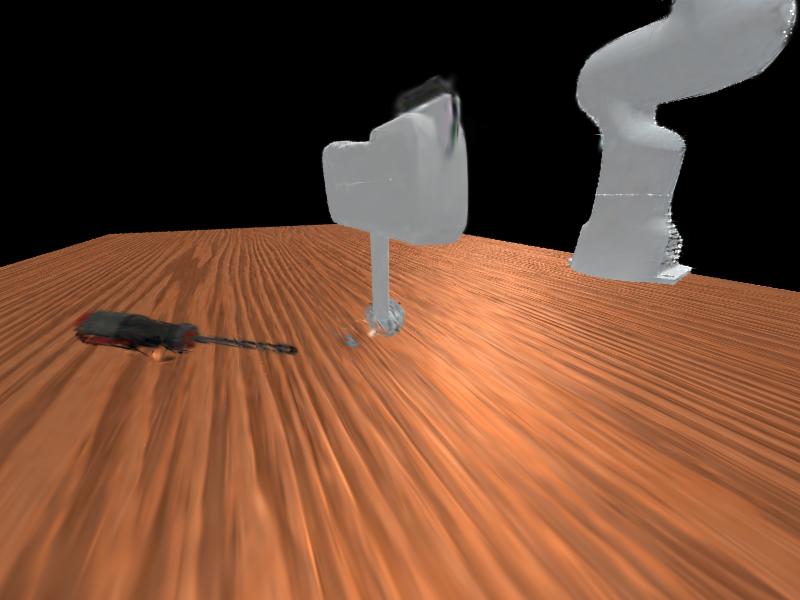}\\
        \end{tabular}
        \end{minipage}\hfill
    \vspace{0.1ex}
    \begin{minipage}{\linewidth}
        \centering
        \begin{tabular}{cccccc}
        \includegraphics[width=0.2\linewidth]{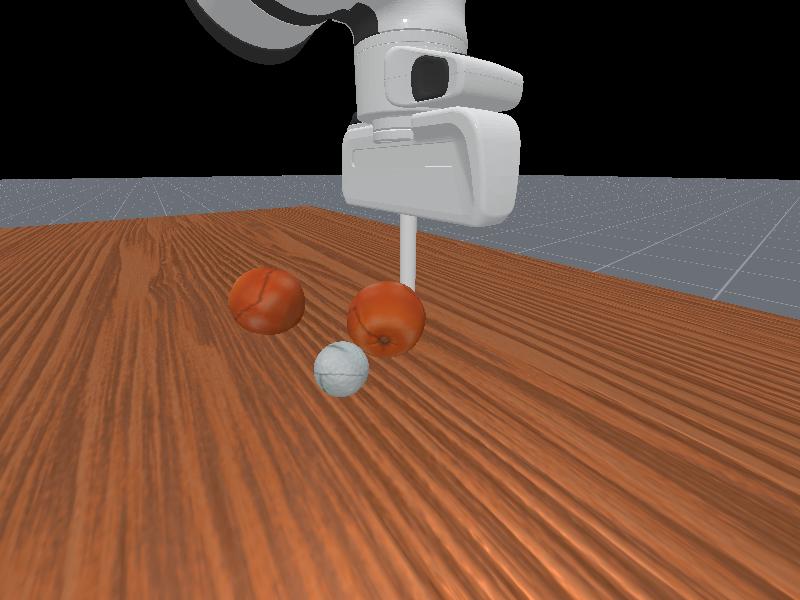} &
        \includegraphics[width=0.2\linewidth]{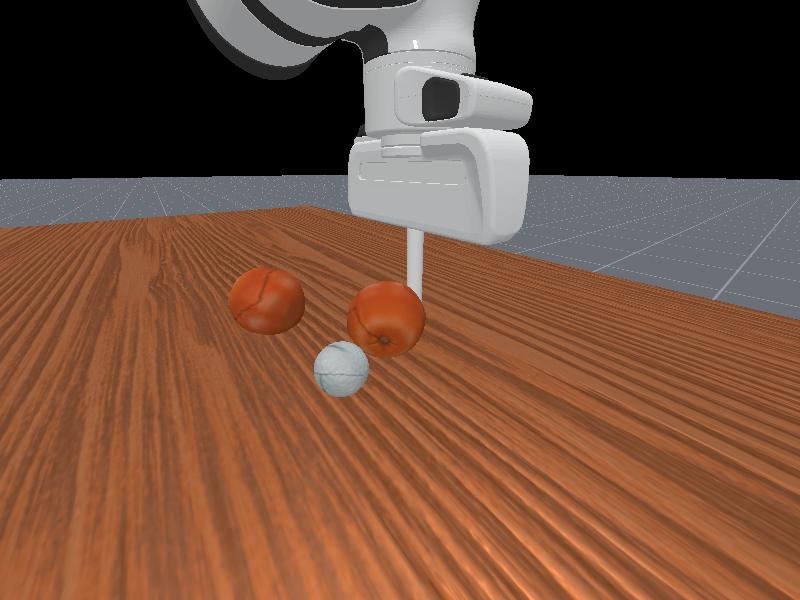} &
        \includegraphics[width=0.2\linewidth]{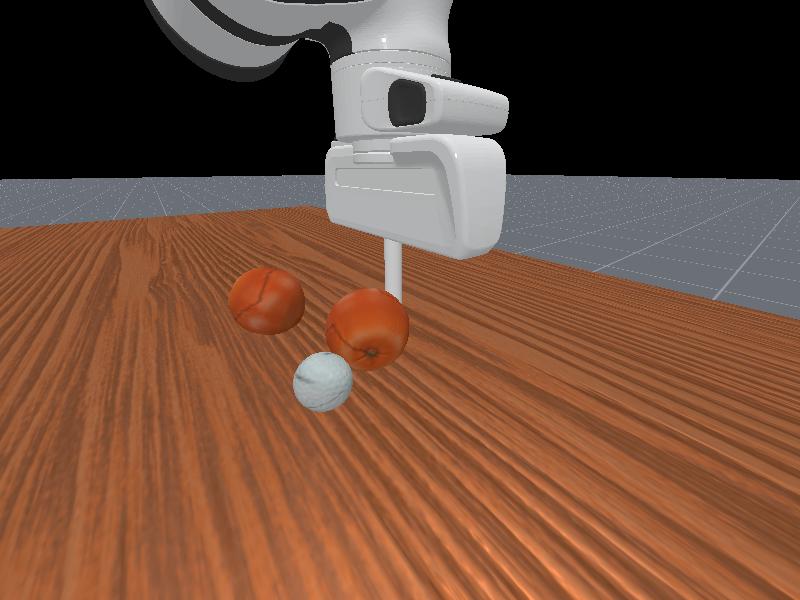} &
        \includegraphics[width=0.2\linewidth]{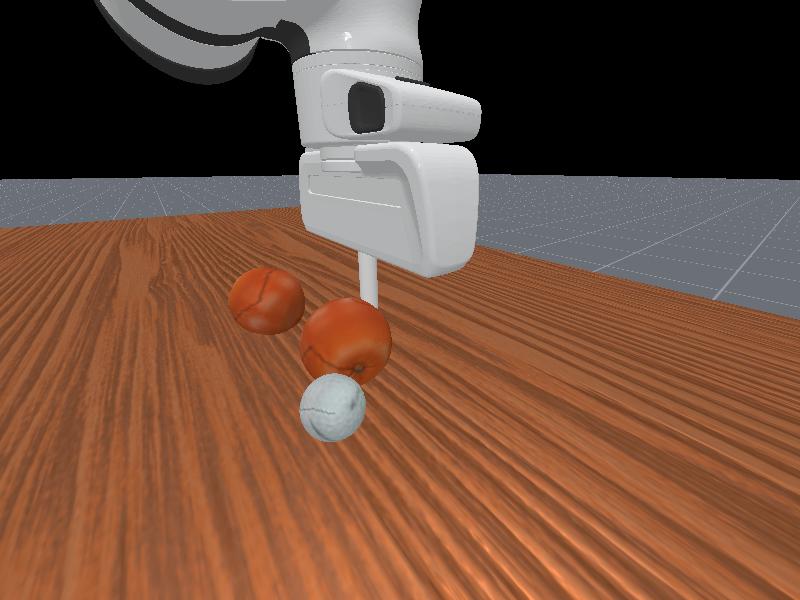} &
        \includegraphics[width=0.2\linewidth]{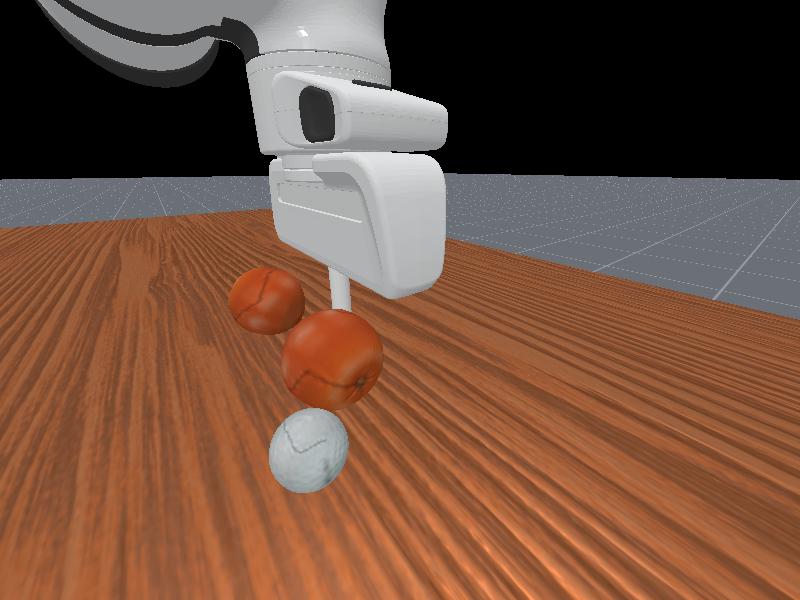}\\
        \includegraphics[width=0.2\linewidth]{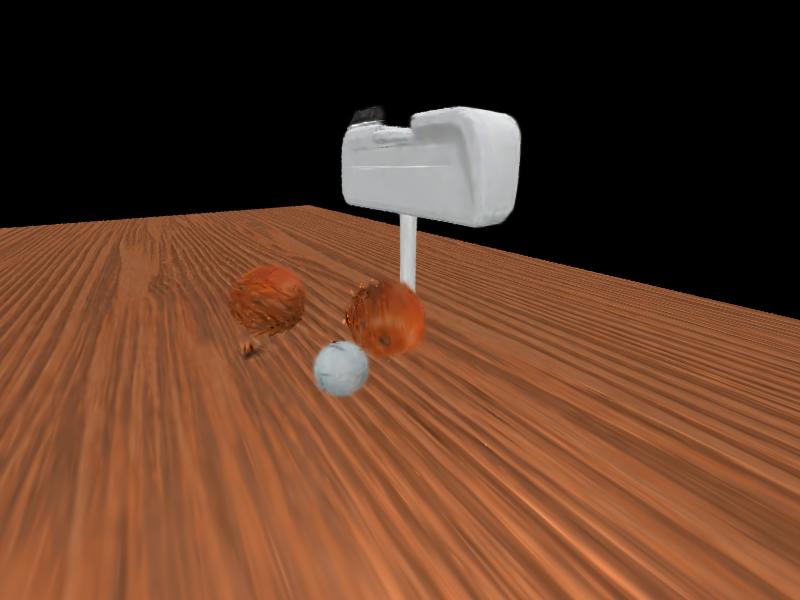} &
        \includegraphics[width=0.2\linewidth]{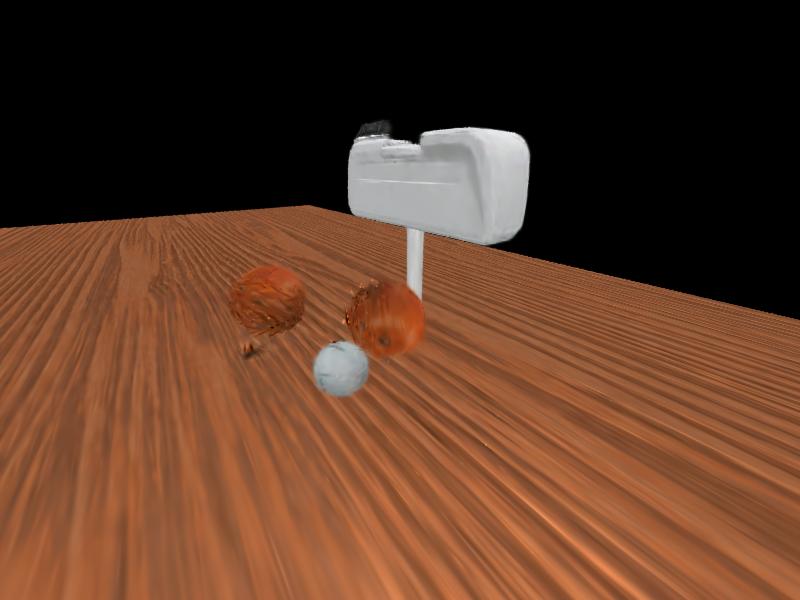} &
        \includegraphics[width=0.2\linewidth]{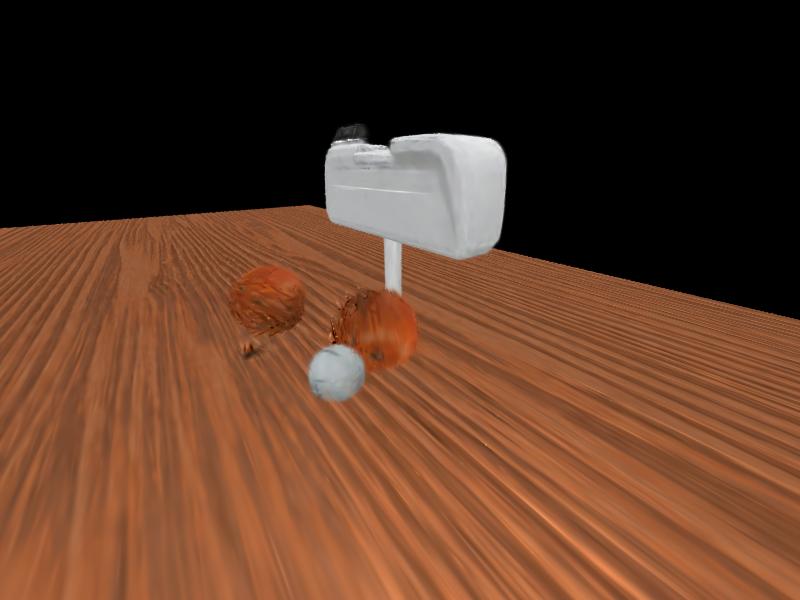} &
        \includegraphics[width=0.2\linewidth]{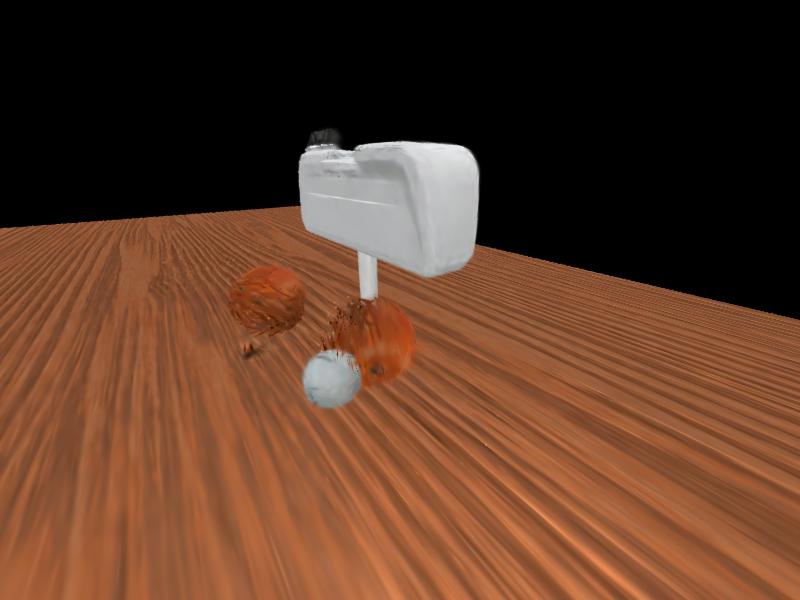} &
        \includegraphics[width=0.2\linewidth]{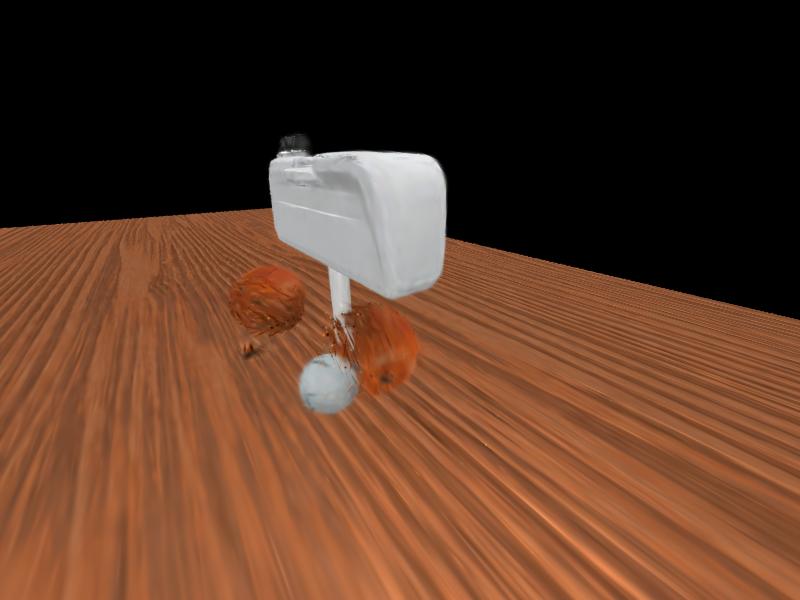}\\
        \end{tabular}
    \end{minipage}\hfill
    \vspace{0.1ex}
    \begin{minipage}{\linewidth}
        \centering
        \begin{tabular}{cccccc}
        \includegraphics[width=0.2\linewidth]{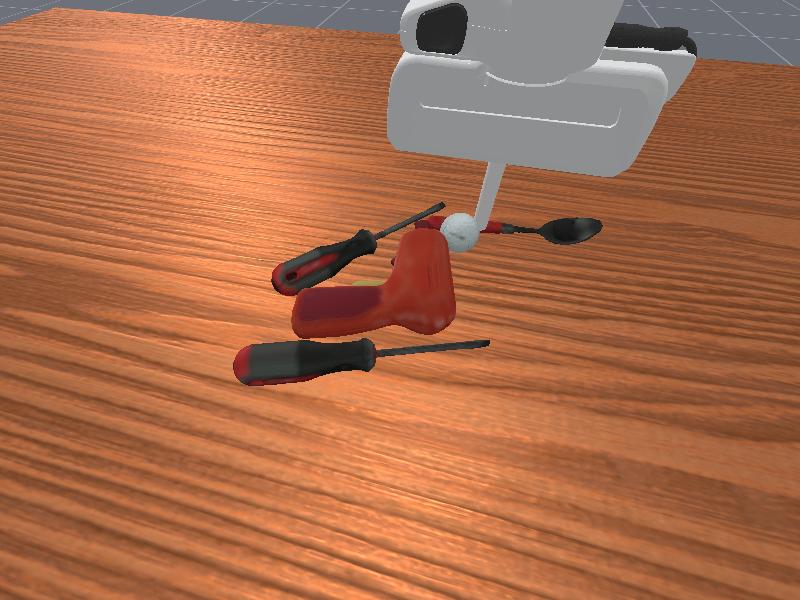} &
        \includegraphics[width=0.2\linewidth]{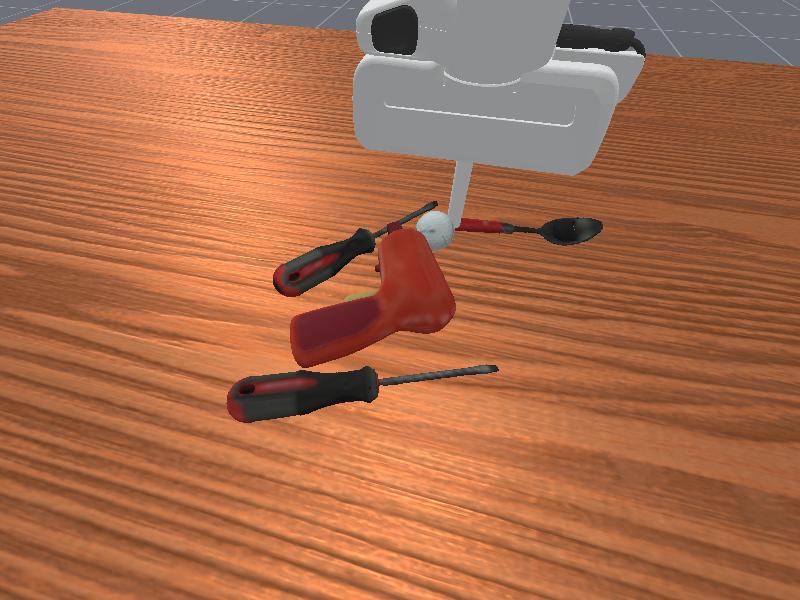} &
        \includegraphics[width=0.2\linewidth]{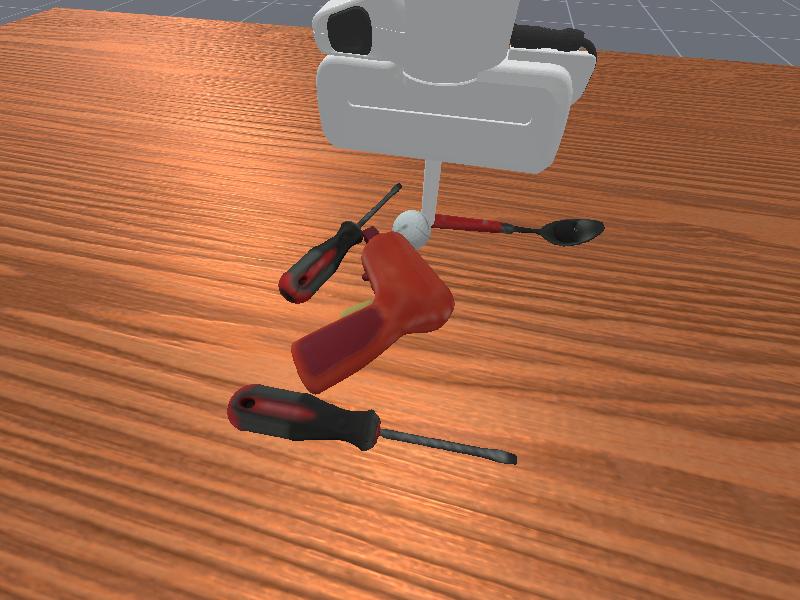} &
        \includegraphics[width=0.2\linewidth]{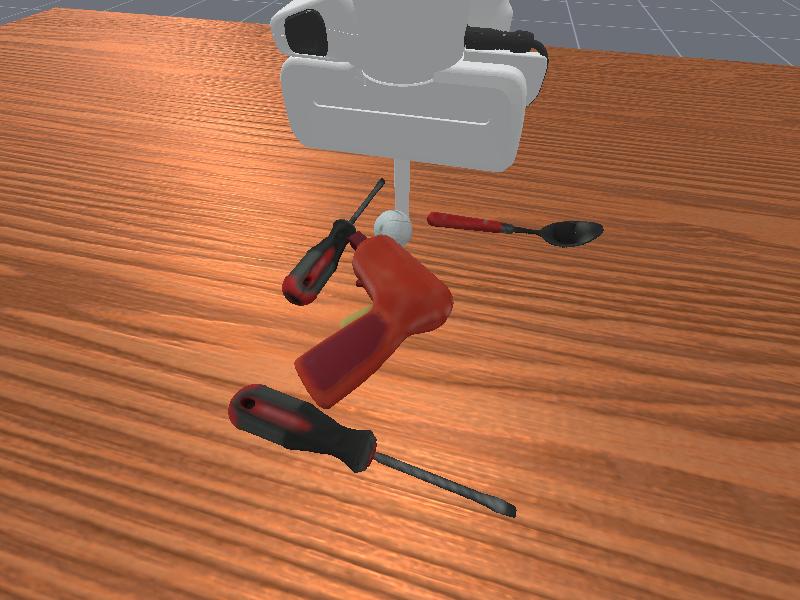} &
        \includegraphics[width=0.2\linewidth]{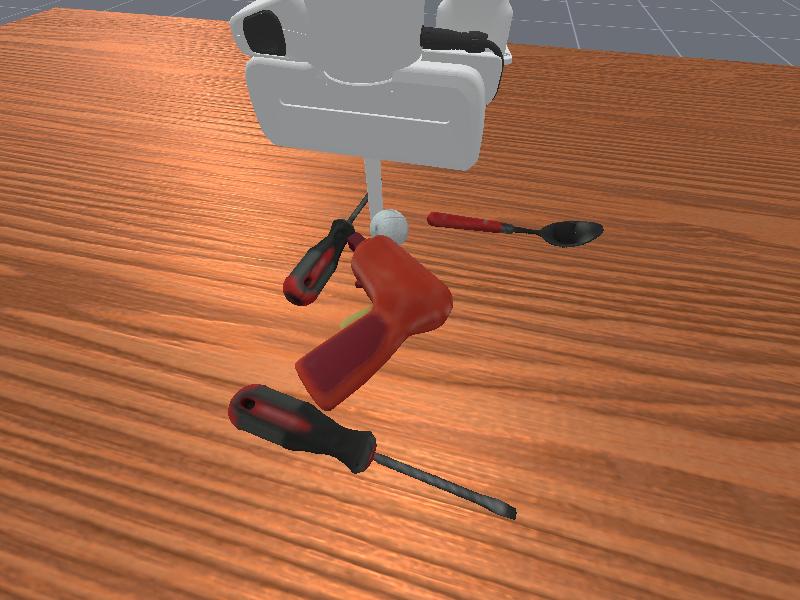}\\
        \includegraphics[width=0.2\linewidth]{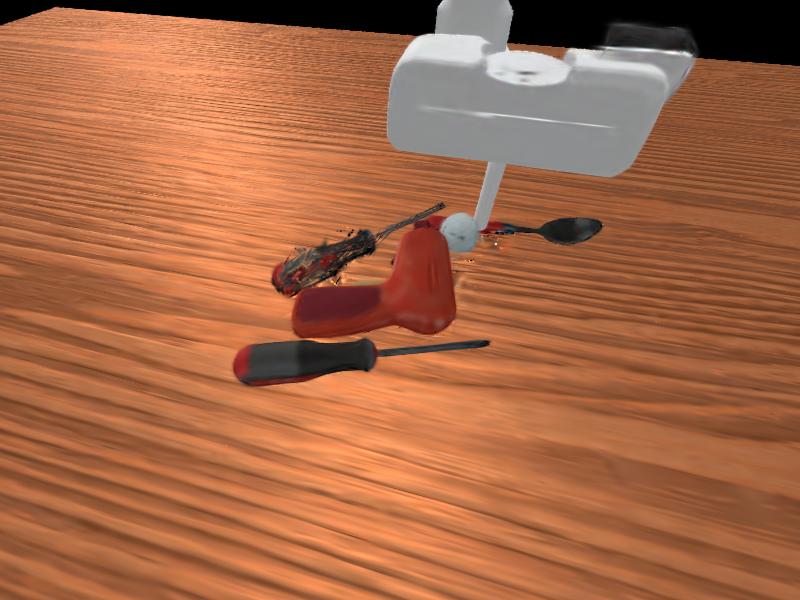} &
        \includegraphics[width=0.2\linewidth]{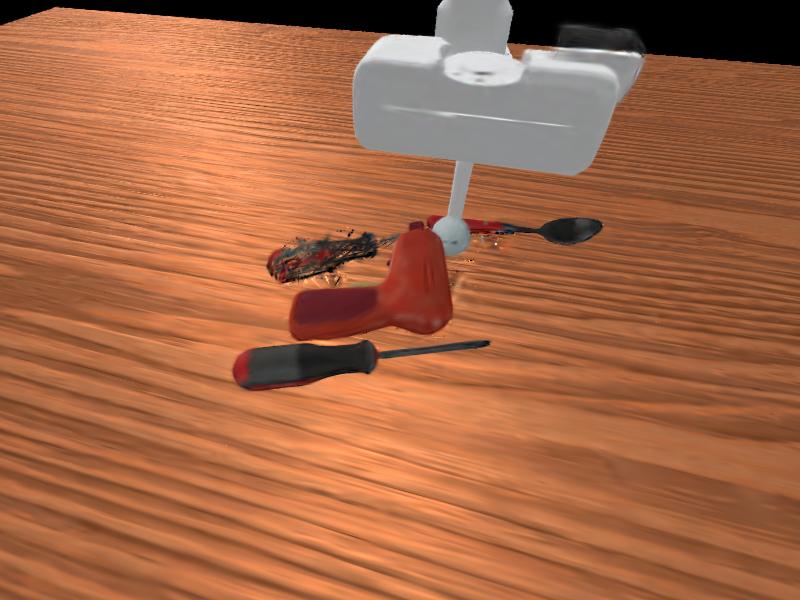} &
        \includegraphics[width=0.2\linewidth]{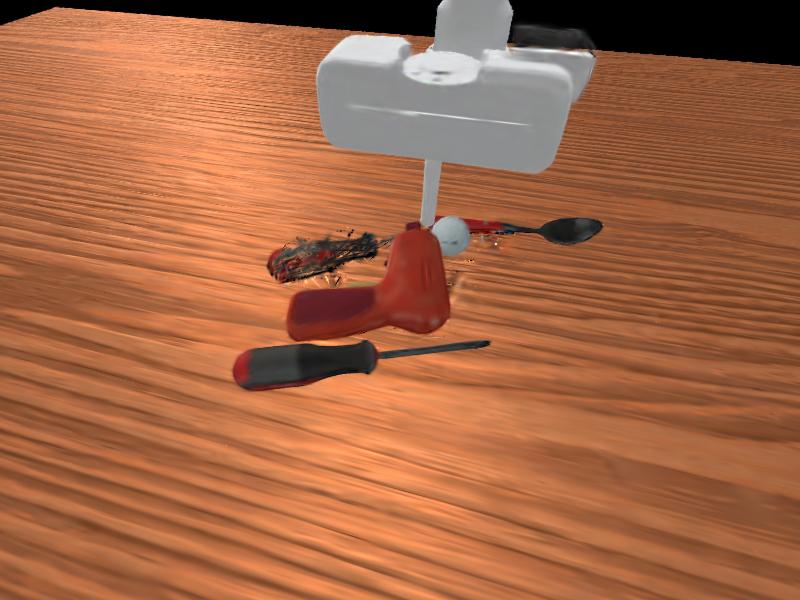} &
        \includegraphics[width=0.2\linewidth]{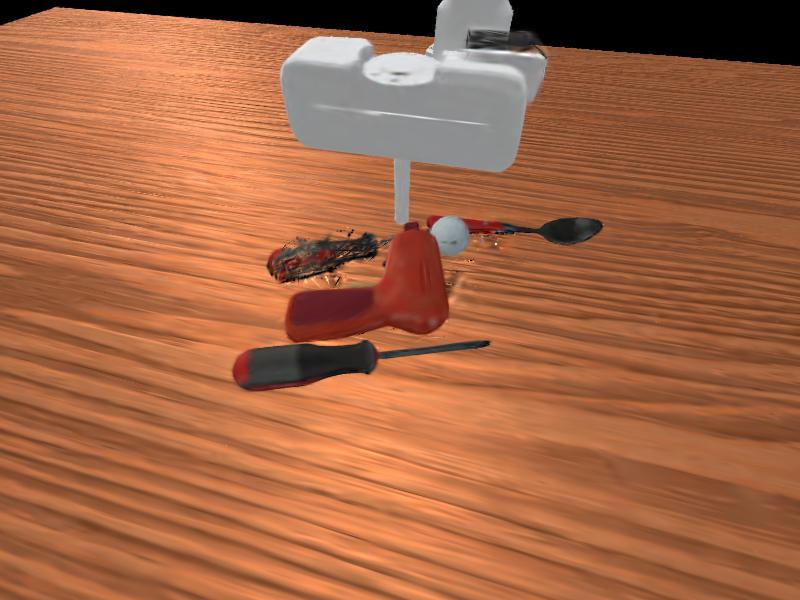} &
        \includegraphics[width=0.2\linewidth]{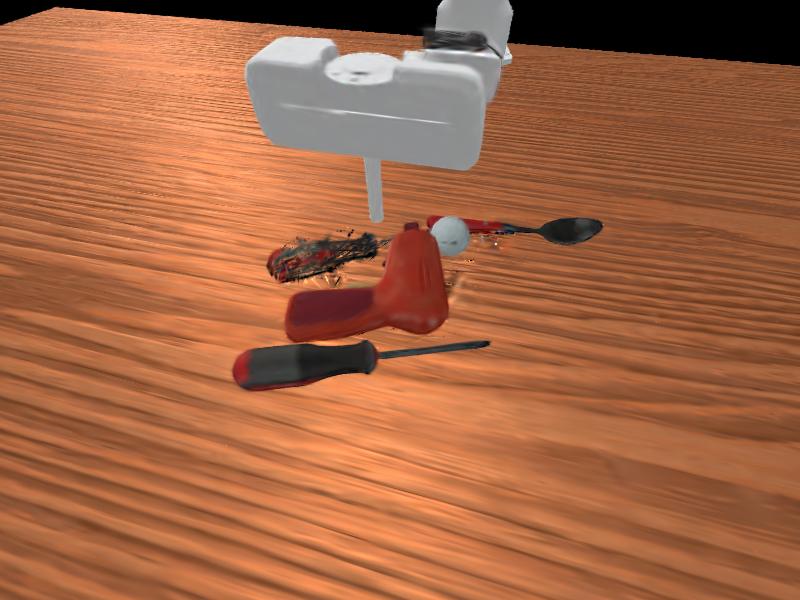}\\
        \end{tabular}
    \end{minipage}\hfill
    \caption{Predicted sequences over horizon $\SI{2.4}{\second}$ (3 model invocations) and ground truth (atop predictions for each sequence) replayed in simulation.
    From top to bottom: smallest, second smallest, median and worst prediction error rank (see text for details on the ranking).
    The object interactions are predicted closely to the ground truth for the two smallest error ranks. 
    For the median and worst ranks, small deviations in the interactions between gripper and objects can cause large deviations in object motion and are harder to predict.
    }
    
    \label{fig:pred-qualitative}
\end{figure}
In \Cref{fig:pred-qualitative}, we show rendered scene representations over time for ground truth and model predictions.
For predictions, we render the Gaussian representation.
We determine the evaluated chunks with 10\% largest changes in accumulated ground-truth position and rotation at 2.4\,s prediction horizon (3 recurrent model calls).
These chunks are ranked according to the sum of position and rotation prediction error.
We visualize the best, second best, median and worst rank example.
In the examples, meaningful interactions can be observed, where for the ones with smallest rank the model predicts motion of a single round object which closely follows the ground truth.
For the median example, the predicted rolling direction of the orange object differs from the actual one and the orange interacts differently with the golf ball.
This might be in parts due to the slightly non-spherical shape of the orange object.
The worst example shows a difficult multi-object interaction setting where a small change in end-effector position can result in contact or non-contact with an object and a large difference in motion.
Videos of the results can be found on the \href{https://embodiedvision.github.io/mro-gwm/}{project website}.

\subsection{Planning Results}

\begin{figure}[t]
\vspace{2mm}
    \centering
    \setlength{\tabcolsep}{0pt}
    \renewcommand{\arraystretch}{0.}
        \begin{subfigure}{\linewidth}
        \centering
        \begin{tabular}{cccccc}
        \includegraphics[width=0.2\linewidth]{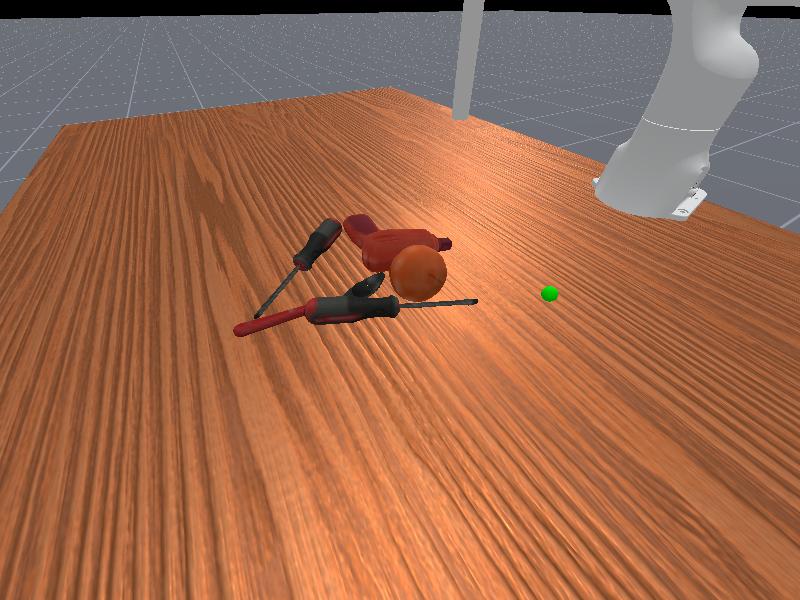}%
        \includegraphics[width=0.2\linewidth]{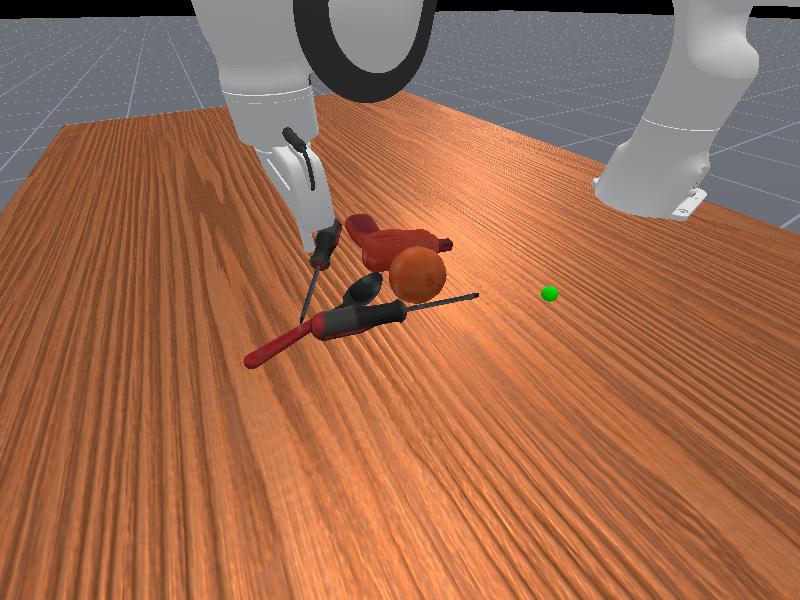}%
        \includegraphics[width=0.2\linewidth]{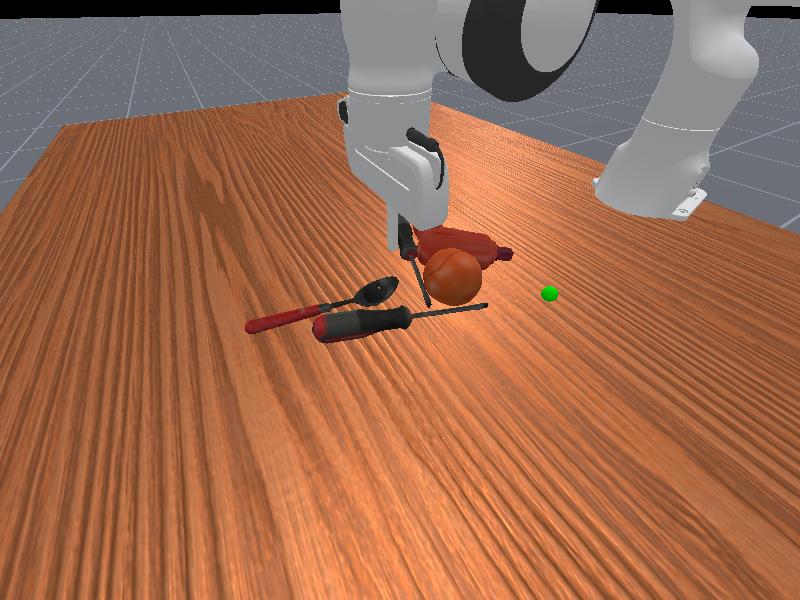}%
        \includegraphics[width=0.2\linewidth]{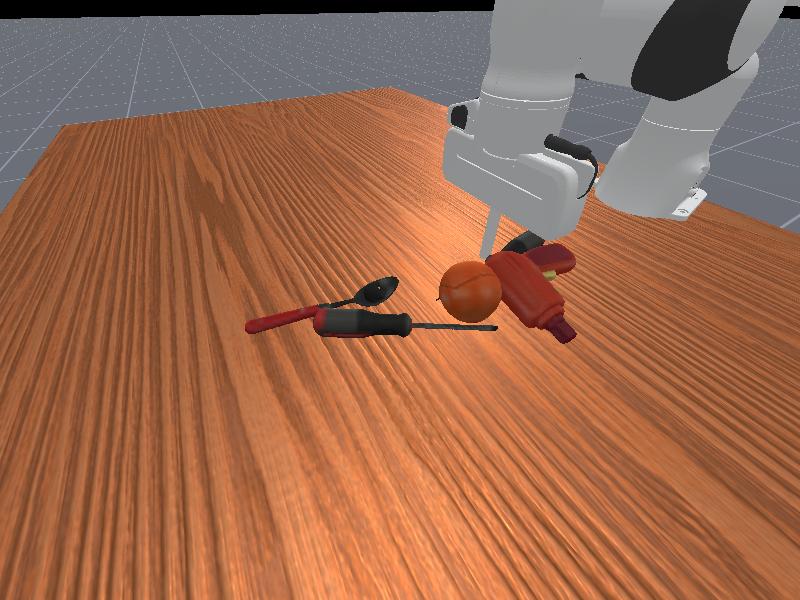}%
        \includegraphics[width=0.2\linewidth]{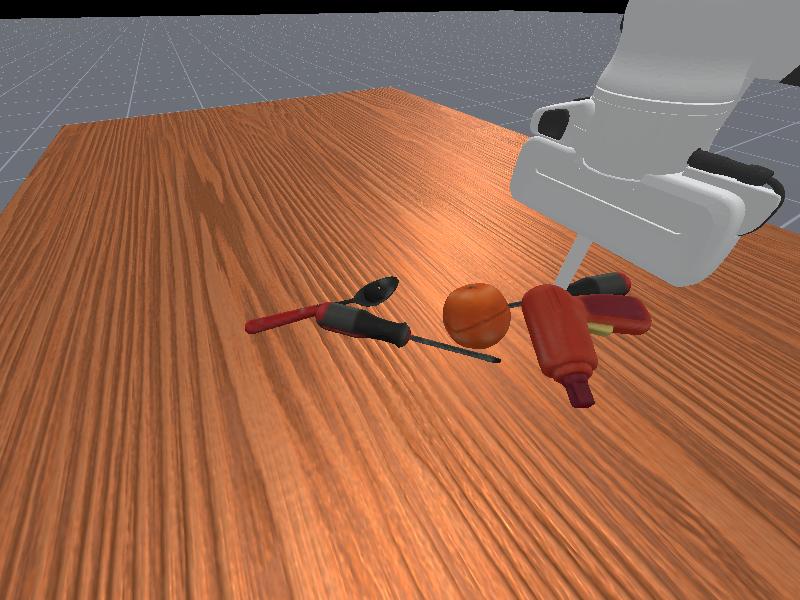}%
        \end{tabular}
    \end{subfigure}
    \begin{subfigure}{\linewidth}
        \centering
        \begin{tabular}{cccccc}
        \includegraphics[width=0.2\linewidth]{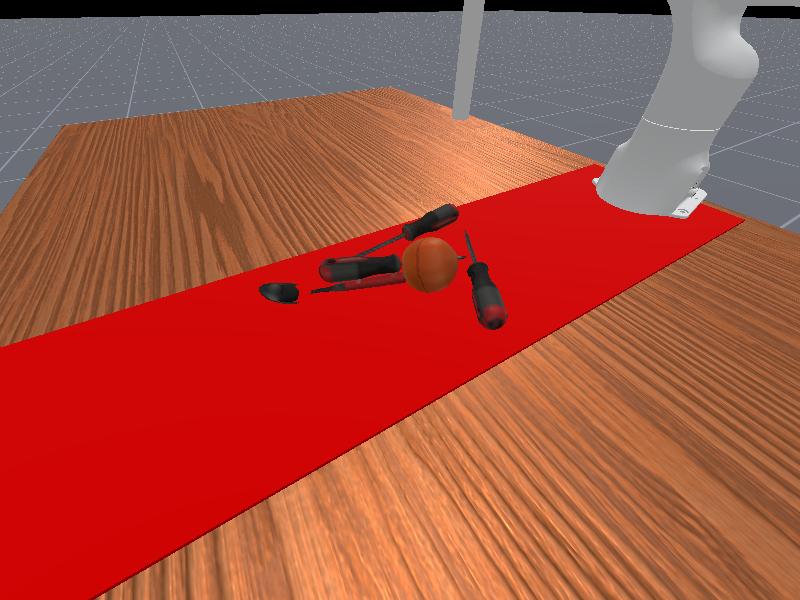}%
        \includegraphics[width=0.2\linewidth]{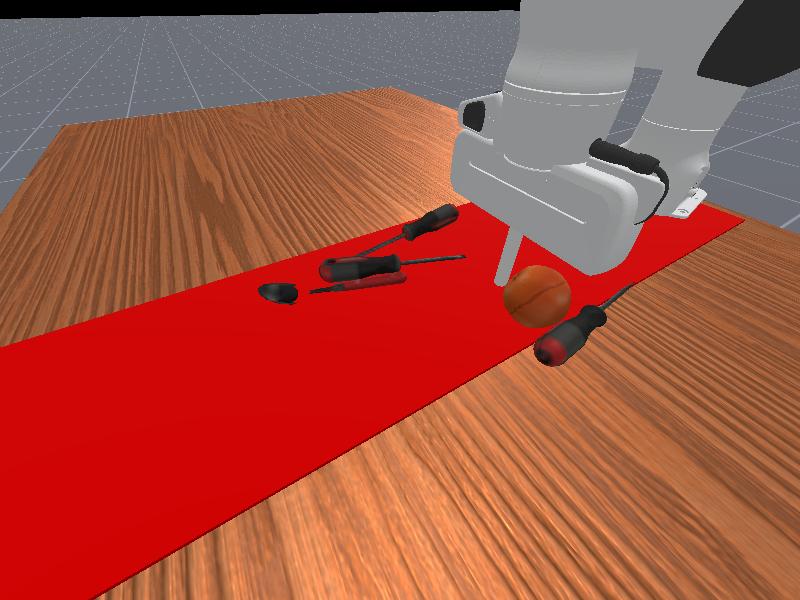}%
        \includegraphics[width=0.2\linewidth]{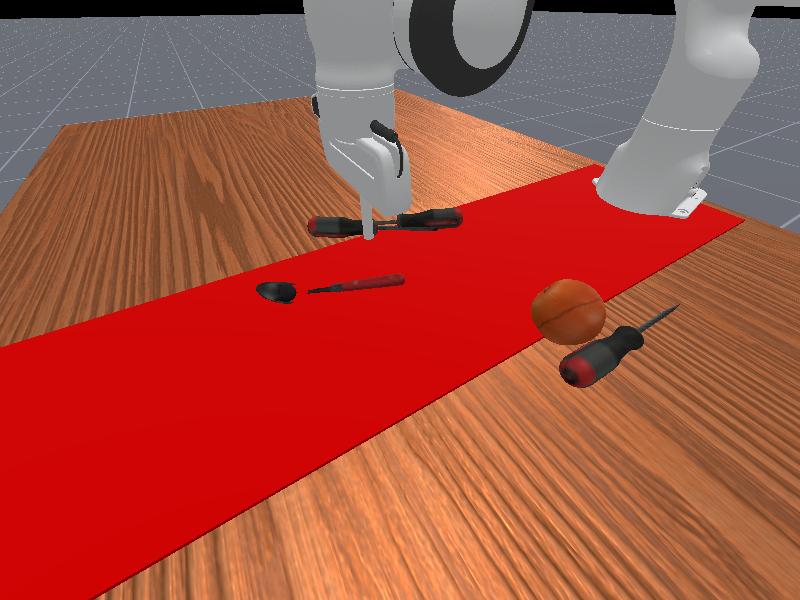}%
        \includegraphics[width=0.2\linewidth]{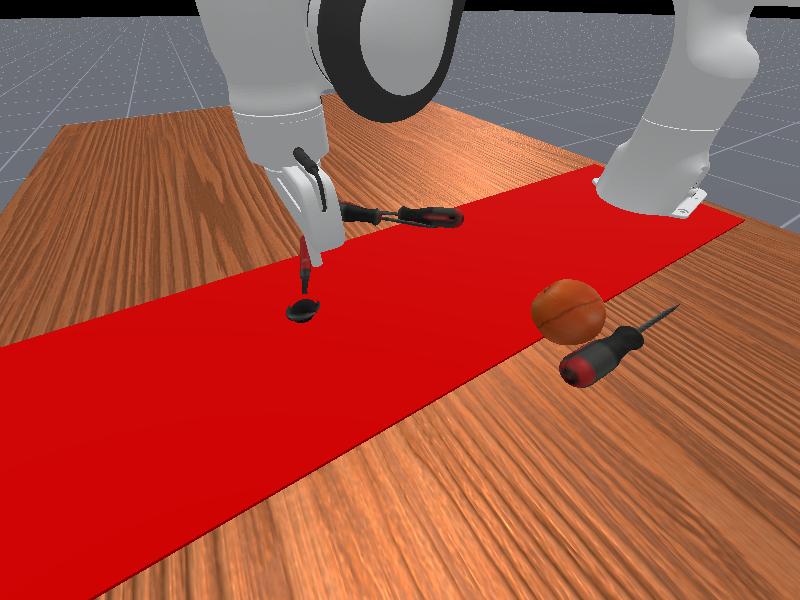}%
        \includegraphics[width=0.2\linewidth]{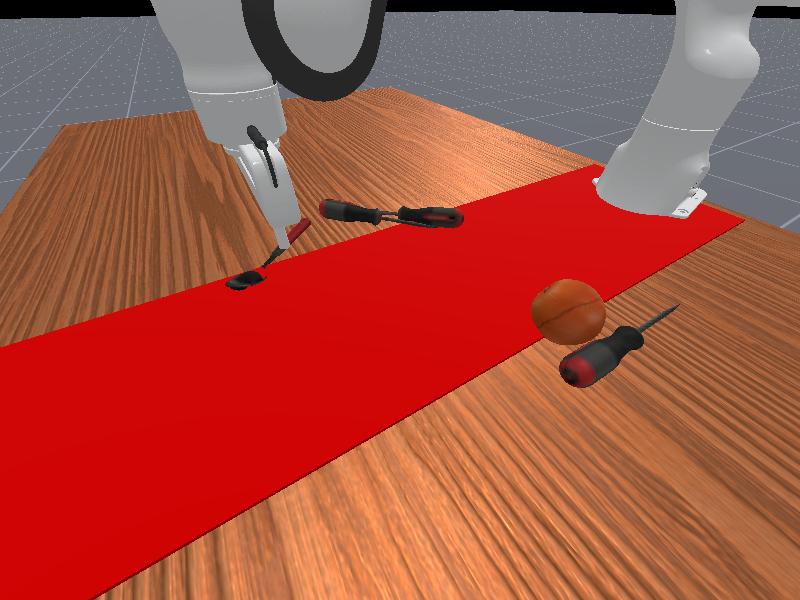}%
        \end{tabular}
    \end{subfigure}
    \caption{Examples for MPC with 5 objects and largest initial objective value. Top: task 1, push object (left screwdriver) to position (green sphere). Bottom: task 2, clear middle (red area). The tasks are successfully solved.
    }
    \label{fig:plan-examples}
\end{figure}

\begin{table}[t]
    \vspace{2mm}
    \centering
    \footnotesize
    \setlength{\tabcolsep}{6pt}
    \begin{tabular}{lccccccc}
    \toprule
        task & \makecell{$t_\text{succ}$} & \makecell{SR} & \makecell{SR\\AUC} & \makecell{$d_\text{goal}$\\ final}  & \makecell{$d_\text{goal}$\\ initial}\\
        \midrule
        1. position object  & 18.5 & 0.79 & 0.85 & 2.57 & 14.29\\
        2. clear middle, all & 24.8 & 0.74 & 0.72 & 3.19 & 30.23\\
        \phantom{2.} cl. mid., all, heuristic & 26.0 & 0.14 & 0.15 & 23.65 & 30.23\\
        \phantom{2.} clear middle, per obj & & 0.88 & 0.83 & & \\
    \bottomrule
    \end{tabular}
    \caption{Planning performance on different tasks where our model is used in a MPC setting. 
        All metrics are computed by taking the mean over all episodes for one training model seed.
        Final goal distance is measured at success or time out.
        Then, the mean over 5 model training seeds is computed and shown here.
        A majority of runs is successful and error metrics are reduced from their initial values.
    }
    \label{tab:plan-results}
\end{table}

We provide planning results in \Cref{tab:plan-results}.
Our approach achieves success rates (SR) of 0.79 and 0.74 in the first and second task, respectively.
When measuring for success for individual objects in the second task, our approach achieves a higher success rate of 0.88.
Goal distances for both tasks are strongly reduced compared to their values at the episode start.
For task 2, we also show that the heuristic policy that moves towards the object with the highest goal distance achieves a success rate of 0.14 and only slightly reduces goal distance.
This demonstrates that the capability to predict the consequences of actions and planning with the world model is important.
In \Cref{fig:plan-examples}, we visualize successful episodes for both tasks on scenes of 5 objects with the largest initial objective value.
In the task 1 example, the left target screwdriver is successfully moved to the goal.
The task 2 example shows that the robot pushes two objects together twice initially.
While it incorrectly moves the remaining spoon from the side first, it finds a better mode by pushing differently and achieves success.
We perform visual inspection of failed episodes for both tasks on scenes with five objects (for one model training seed) and identify the following typical failure cases:
Multiple episodes of task 2 time-out due to the robot switching several times between approaching different target objects or getting stuck in the center of an object cluster with no clear incentive which object to interact with next.
Such behavior could be alleviated by a more intricate design of the cost functions.
In some failure cases of both tasks the robot struggles to manipulate the screwdriver, the spoon and the power drill. Elongated objects seem difficult to predict, presumably since the center of mass strongly affects rotation.

\subsection{Joint Tracking and Prediction Results}

\begin{table}[t]
    \vspace{2mm}
    \centering
    \footnotesize
    \setlength{\tabcolsep}{5pt}
    \begin{tabular}{lcccccc}
    \toprule
     \multirow{2}{*}{} & \multicolumn{2}{c}{pos err} & \multicolumn{2}{c}{rot err} & &   \\
             & abs & rel & abs & rel &  ADD & ADD-S \\
    \midrule
    tracking   & 0.35 &     & 5.09 & &   0.57 & 0.29\\
    pred., gt input & 1.77 & 1.77 & 13.51 & 13.51 & 2.35 & 1.38 \\
    pred., tracked input  & 2.24 & 2.10 & 19.44 & 15.73 & 3.05 & 1.69\\
    \bottomrule
    \end{tabular}
    \caption{Mean errors for tracking, computed over moving chunks only. Position error and ADD(-S) in \unit{\centi\meter}, rotation error in \unit{deg}. Averaged over 5 model training seeds. Tracking introduces estimation errors which also affect predictions.
    }
    \label{tab:track-results}
\end{table}

\begin{figure}[t]
    \centering
    \vspace{2mm}
    \setlength{\tabcolsep}{0pt}
    \renewcommand{\arraystretch}{0.}
    \begin{minipage}{\linewidth}
        \centering
        \begin{tabular}{ccccc}
        \includegraphics[width=0.25\linewidth]{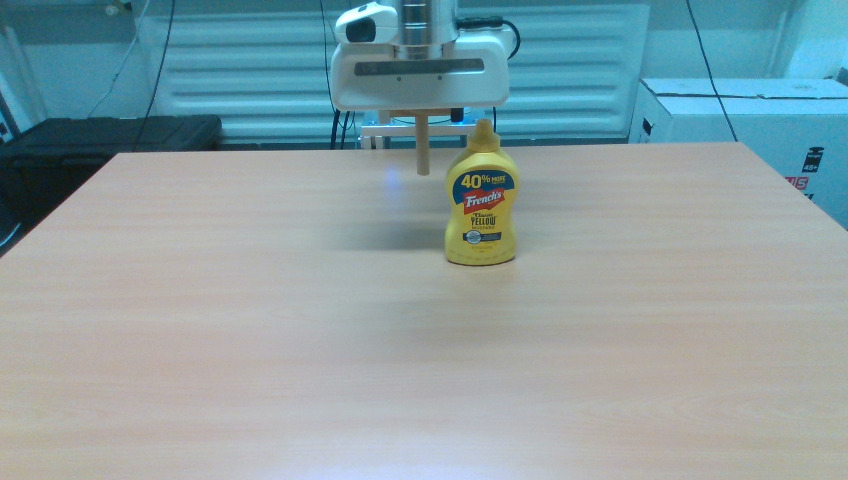} &
        \includegraphics[width=0.25\linewidth]{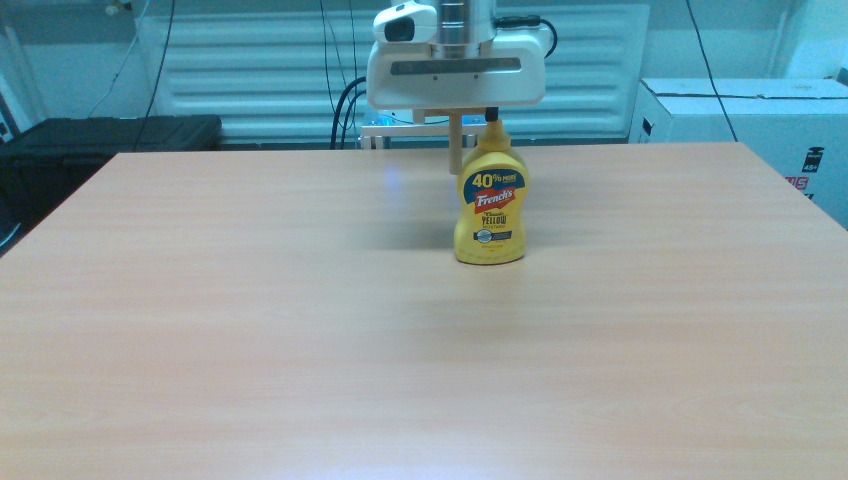} &
        \includegraphics[width=0.25\linewidth]{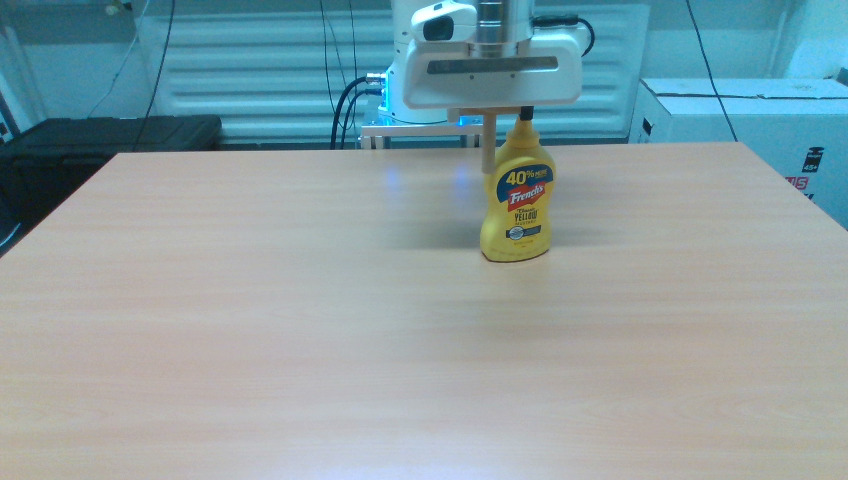} &
        \includegraphics[width=0.25\linewidth]{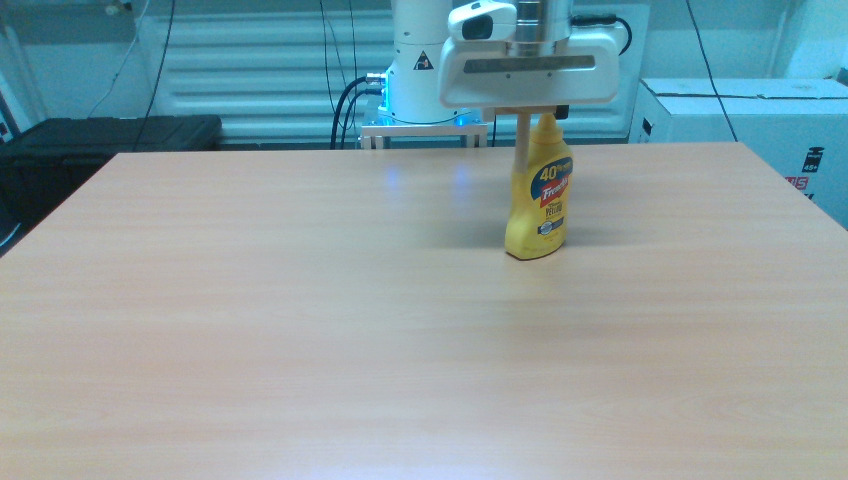}\\
        \includegraphics[width=0.25\linewidth]{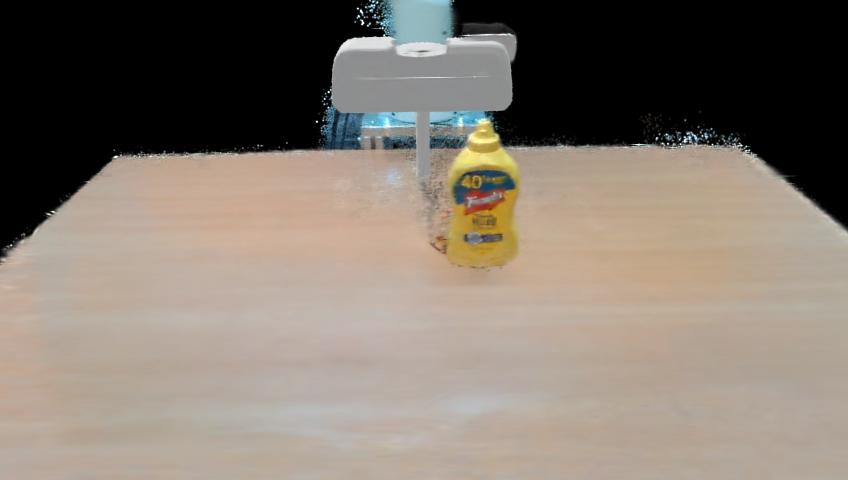} &
        \includegraphics[width=0.25\linewidth]{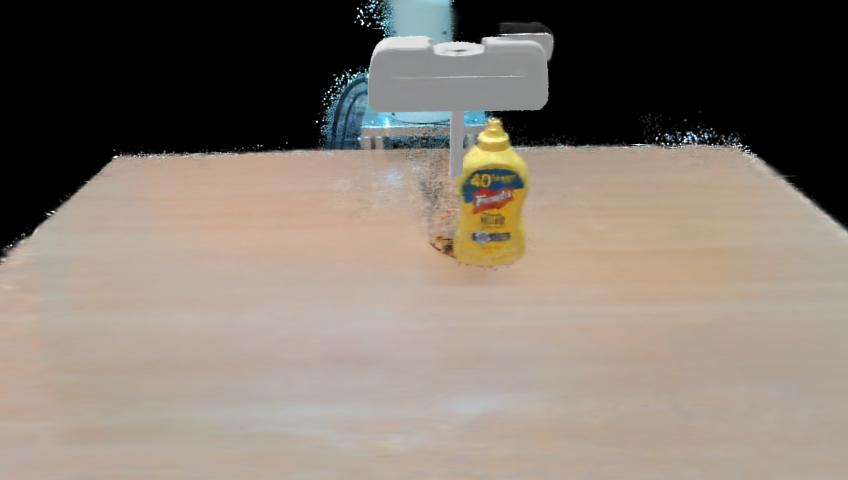} &
        \includegraphics[width=0.25\linewidth]{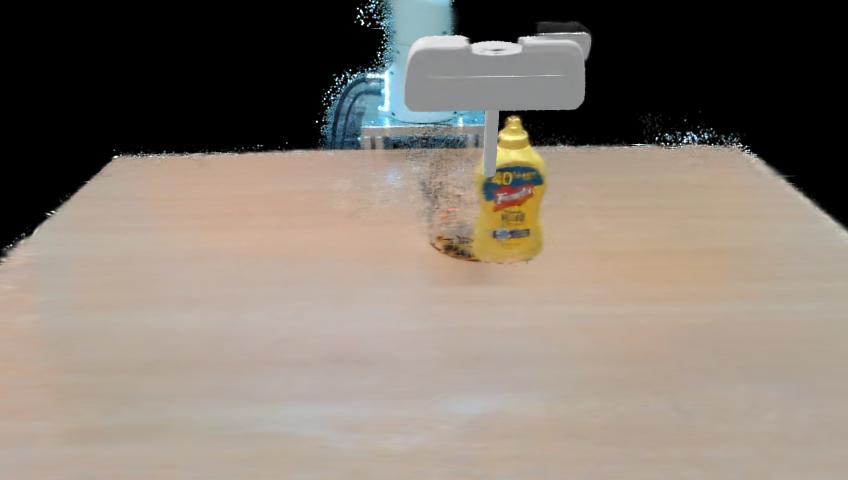} &
        \includegraphics[width=0.25\linewidth]{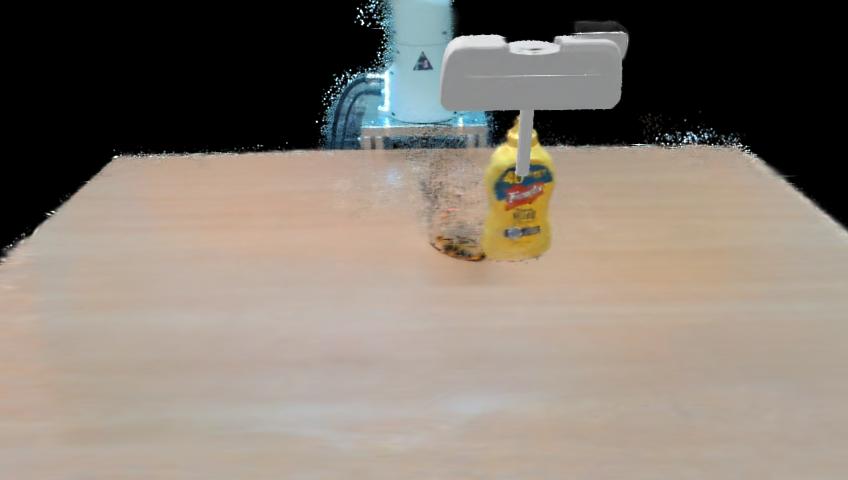}\\
        \end{tabular}
    \end{minipage}\hfill
    \vspace{0.5ex}
    \begin{minipage}{\linewidth}
        \centering
        \begin{tabular}{ccccc}
        \includegraphics[width=0.25\linewidth]{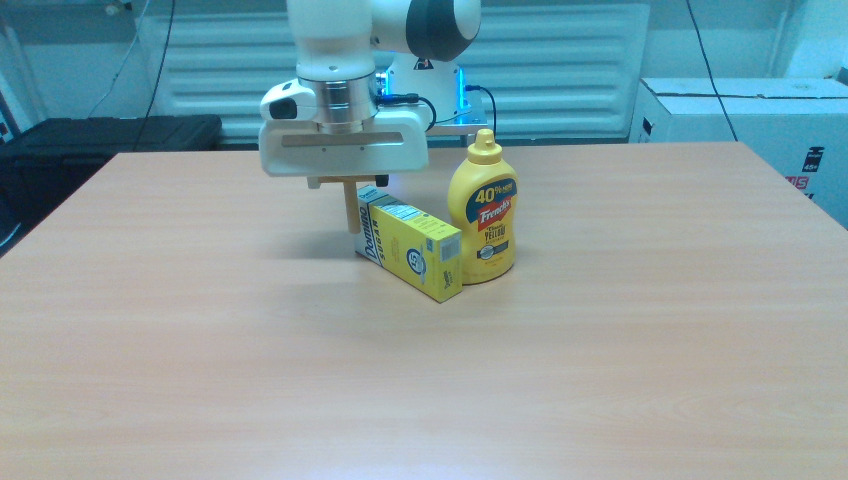} &
        \includegraphics[width=0.25\linewidth]{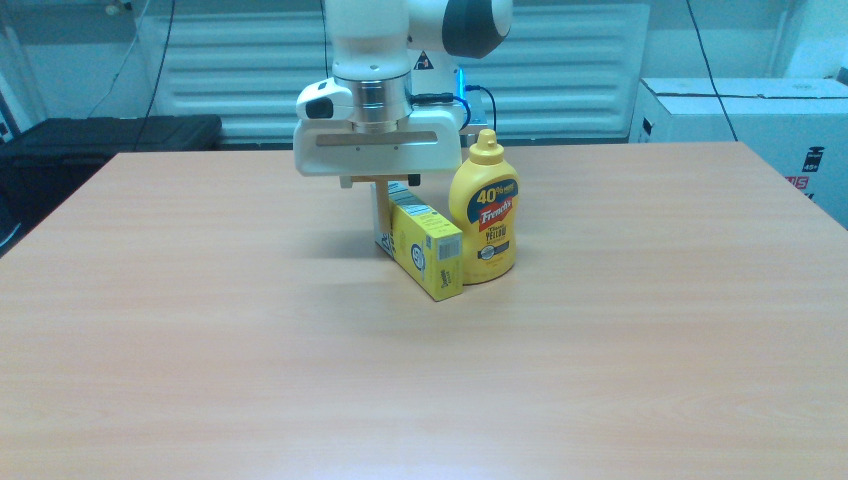} &
        \includegraphics[width=0.25\linewidth]{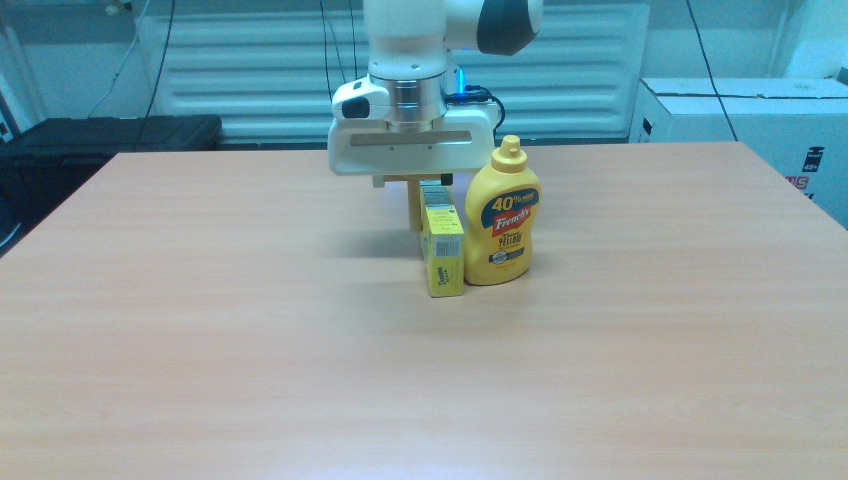} &
        \includegraphics[width=0.25\linewidth]{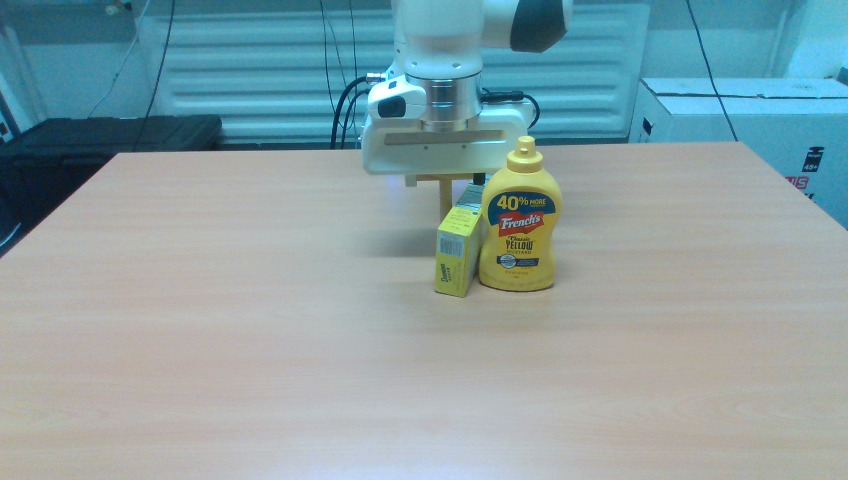}\\
        \includegraphics[width=0.25\linewidth]{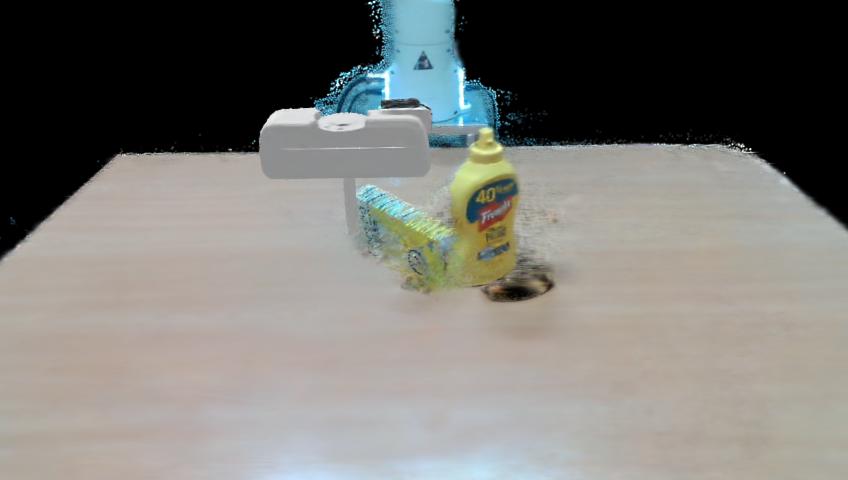} &
        \includegraphics[width=0.25\linewidth]{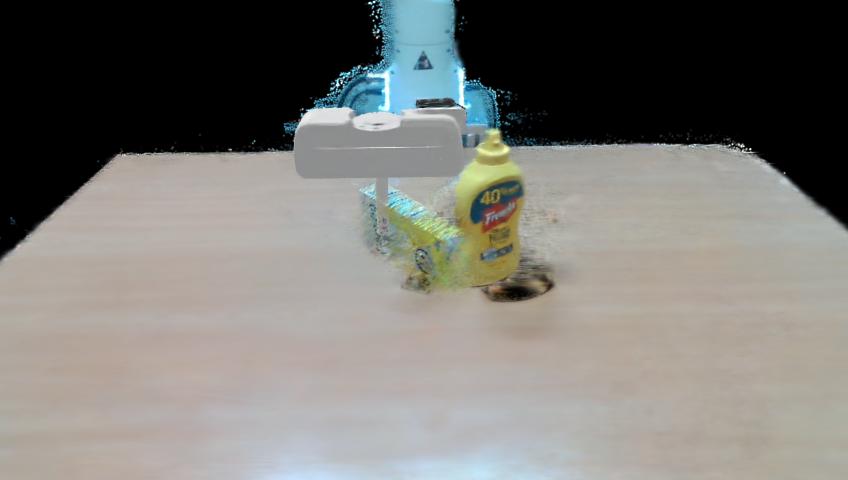} &
        \includegraphics[width=0.25\linewidth]{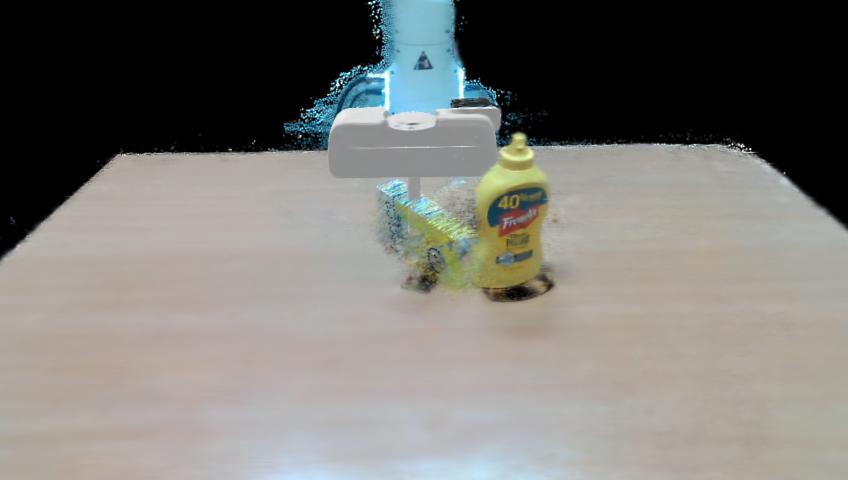} &
        \includegraphics[width=0.25\linewidth]{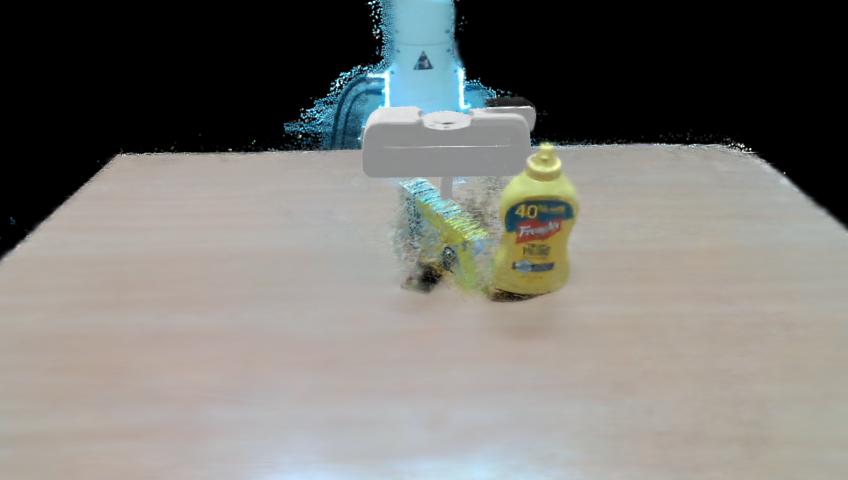}\\
        \end{tabular}
    \end{minipage}\hfill
    \caption{Qualitative real-world tracking and prediction results. The objects are tracked until the first frame and our model is used to predict interactions of objects and end-effector.
    }
    
    \label{fig:track-qualitative}
\end{figure}

\addtolength{\textheight}{-1cm}   

In \Cref{tab:track-results}, we provide results for object tracking in images and dynamics prediction based on the estimated object poses.
We evaluate chunks classified as moving at a prediction horizon of \SI{2.4}{\second}, i.e., 3 recurrent dynamics model calls.
Tracking error is evaluated at the last time step of the input history for each chunk.
Notably, ADD-S~\cite{xiang2018_posecnn} is smaller than ADD, which hints at discrete rotational symmetries of the objects that get confused by the pose estimation.
Prediction error is evaluated at the end of the prediction horizon for ground-truth and estimated trajectory inputs.
Relative errors are computed with respect to the last history step.
One can observe that the absolute prediction errors on tracked input are higher than for ground truth input and slightly higher than adding the errors from the tracking.
Expectedly, dynamics prediction performs worse on tracked trajectories, presumably due to tracking errors and trajectory noise not observed in the training data.
In \Cref{fig:pred-qualitative}, we demonstrate qualitative results for tracking and prediction in two real-world recordings, in which a Panda robot pushes objects in predefined end-effector motions.
The scene is captured using an Intel RealSense D435 camera and camera poses are obtained using~\cite{orbslam}.
In these examples, our approach tracks object pose well and predicts plausible interactions between objects and the end effector initially.
At later steps in the second scene, the model misses the interaction between box and end effector, presumably as prediction gets out of distribution due to prediction errors, e.g., objects and end effector penetrate, or insufficient training data which could be addressed in future work by noise simulation or domain randomization.

\section{Conclusions}
We present MRO-GWM, a novel action-conditional world model that operates on an object-centric Gaussian splatting representation and predicts rigid body movement.
The Gaussians are compressed in anchors which are represented in the object frames and are rigidly transformed to condition the model on a history of object poses.
We further propose a transformer architecture that operates on this representation and uses temporal and spatial attention layers as well as a new temporal-spatial attention layer.
We evaluate our model on simulated tabletop scenes in which up to five typical household objects are closely placed next to each other and interaction is caused by a robot end effector.
We compare prediction performance of several model variants and ablations.
We also demonstrate the performance of our approach for image-based object tracking and prediction in simulated and real scenes and that model-predictive control with our model can successfully tackle two non-prehensile manipulation tasks in simulation in several multi-object configurations.
A limitation of our approach is that Gaussian reconstruction precedes tracking or planning in separate phases. 
Incremental reconstruction is an interesting direction for future research.
Future research could also investigate improving sim2real transfer by noise modeling and domain randomization.
Improving efficiency of model or planning for real-time control is also interesting for future work.

\section*{Acknowledgment}
We thank Marlene Hermelink for her support in software development of the data preprocessing pipeline for the real-world experiments.



{
    \bibliographystyle{IEEEtran}
    \bibliography{main}
}

\end{document}